\documentclass{article}

\usepackage{PRIMEarxiv}
\usepackage[authoryear]{natbib}
\usepackage{amsmath}
\usepackage{amsfonts}
\usepackage{float}
\usepackage{pifont}
\usepackage[no-math]{fontspec}
\usepackage{tcolorbox}
\usepackage{etoolbox}
\usepackage{xspace}
\usepackage{hyperref}
\usepackage{url}
\usepackage{subcaption}
\usepackage{booktabs}
\usepackage{nicefrac}
\usepackage{multirow}
\usepackage{microtype}
\usepackage{enumitem}
\usepackage{lipsum}
\usepackage{fancyhdr}
\usepackage{graphicx}

\usepackage[english,bidi=bidi]{babel}
\babelprovide[import]{arabic}

\babelfont[arabic]{rm}[Path=fonts/]{Amiri-Regular.ttf}
\babelfont[arabic]{sf}[Path=fonts/]{Amiri-Regular.ttf}
\babelfont[arabic]{tt}[Path=fonts/]{Amiri-Regular.ttf}

\newcommand{\textarabic}[1]{\foreignlanguage{arabic}{#1}}

\def\tsc#1{\csdef{#1}{\textsc{\lowercase{#1}}\xspace}}
\tsc{WGM}
\tsc{QE}

\graphicspath{{media/}}

\title{Towards Figurative Language Understanding in Moroccan Darija: An Evaluation of Large Language Models}
\author{
\textbf{Jihad R'baiti}$^{1,2}$,
\textbf{Youssef Hmamouche}$^{1}$,
\textbf{Emmanuele Chersoni}$^{2,3}$,\\
\textbf{Chu-Ren Huang}$^{2}$,
\textbf{Amal El Fallah Seghrouchni}$^{1,4}$\\[0.5em]
$^{1}$International Artificial Intelligence Center of Morocco,\\
Mohammed VI Polytechnic University, Morocco.\\
$^{2}$Department of Language Science and Technology,\\
The Hong Kong Polytechnic University, Hong Kong.\\
$^{3}$Division of Artificial Intelligence and the Humanities,\\
The Hong Kong Polytechnic University, Hong Kong.\\
$^{4}$LIP6, UMR 7606, CNRS, Sorbonne University, Paris, France
}
\begin{document}
\maketitle

\begin{abstract}
Figurative language remains a major challenge for Machine Translation (MT) systems, as the intended meaning of such expressions often goes beyond literal translation. Although recent advances in Large Language Models (LLMs) improved the translation of Arabic dialects, benchmark resources for evaluating figurative language translation in Moroccan Darija remain scarce. To this end, we introduce the Moroccan Darija Figurative Language Corpus (MDFC), a benchmark dataset for MT and multiple-choice question (MCQ) selection covering idioms, metaphors, and proverbs. Using MDFC, we evaluate 19 open- and closed-weight language models and MT systems  at both the expression and contextual levels under zero-shot and six-shot settings. We also conduct layer-wise similarity-based probing to determine how model representations distinguish figurative from literal meanings. Our findings reveal that closed-weight models achieve the strongest translation performance, while the benefits of contextual information vary across models and tasks. Differences between translation performance and MCQ accuracy highlight the distinction between generating an appropriate figurative translation and recognizing the intended meaning among candidate translations.
\end{abstract}

% keywords can be removed

\section{Introduction}
\noindent Human language plays a central role in communication, allowing individuals to convey ideas, express emotions, and share information across cultures. However, sentence interpretation is not limited to the literal sense of words; it is also shaped by cognitive and pragmatic factors, influenced by cultural background and shared knowledge. This is particularly evident in idioms, metaphors, and proverbs, which function as rhetorical devices whose intended meaning is often implicit rather than explicitly stated. Interpreting these expressions thereby requires linguistic knowledge, cultural awareness, and experience. When translating them, the \textit{literal translation} preserves the direct lexical meaning of the source sentence through word-for-word translation~\citep{oh-etal-2026-tug,baziotis-etal-2023-automatic}. In contrast, the \textit{figurative translation} conveys the intended meaning of the expression while accounting for its cultural and contextual background.

\noindent In the context of MT, LLMs are capable of generating fluent and coherent text~\citep{cui-etal-2025-multilingual, naveed2024comprehensiveoverviewlargelanguage, 10.1162_daed_a_01909}, yet they do not fully understand this context-dependent and socio-pragmatic expressions, such as idioms and slang~\citep{bollepally2026llmsinterpretfigurativelanguage}. Instead, they rely on patterns and representations learned from large-scale training data, which allows them to approximate context-sensitive language use, while struggling to fully capture the meaning of culturally rich and figurative usages of language~\citep{sanchez-bayona-agerri-2025-metaphor,attia-etal-2026-beyond}.

\noindent Despite these limitations, the evaluation of LLMs in MT remains centered on standard benchmarks that prioritize high-resource languages and conventional text~\citep{al-ghallabi-etal-2025-fann,rajaee-etal-2025-empirical}. Such benchmarks provide limited insight into how models handle linguistically complex and culturally grounded phenomena in underrepresented languages~\citep{chakrabarty-etal-2022-flute}. This issue is particularly evident for example for Moroccan Darija, a low-resource Arabic dialect characterized by high variability, lack of standardization, and strong cultural specificity~\citep{Skiredj_2025}. 

\noindent The lack of dedicated evaluation resources for the figurative language makes it difficult to assess the behavior of the model and identify failure modes. For instance, the expression `\textarabic{مقطوع من شجرة}' (literally: `cut from a tree') refers to someone with no family or social ties, while `\textarabic{طار ليه الفريخ}' (literally: `the bird flew away from him') is used to describe someone who has lost their mind or become irrational. Such examples highlight the gap between surface form and intended meaning, reflecting the importance of more targeted evaluation approaches and more robust models capable of handling such figurative expressions.

\noindent Existing benchmarks do not make it possible to classify fluent translation from genuine preservation of figurative meaning, particularly for culturally specific Darija expressions. It therefore remains uncertain whether model improvements, greater scale, or access to contextual information translate into a stronger ability to recover intended meanings rather than reproduce literal lexical content. Addressing this question requires targeted data, comparisons across model families and sizes, and complementary evaluation methods that capture both general translation quality and figurative adequacy.

\noindent To systematically examine both model capability and the methodological factors that shape its evaluation, we address the following research questions:
 How accurately can LLMs translate idiomatic, metaphorical, and proverbial expressions from Moroccan Darija into English while preserving their intended meaning, and to what extent do they rely on literal translation? (\textbf{RQ1})
To what extent does providing contextual information affect models' ability to preserve the intended meaning of Moroccan Darija figurative expressions? (\textbf{RQ2}) 
 What types of translation errors characterize the outputs of the evaluated models, and how do these error patterns differ across selected models? (\textbf{RQ3})
How does model parameter size influence translation performance? (\textbf{RQ4})\\
\noindent The main contributions of this work are as follows:
\begin{itemize}
    \item[-] The introduction of MDFC, the first benchmark specifically designed for Moroccan Darija figurative language translation, comprising expression- and context-based data in which source sentences are paired with literal and figuratively appropriate English translations.

     \item[-] A multidimensional evaluation framework that examines figurative translation generation and interpretation through complementary automatic and LLM-based metrics, multiple-choice evaluation under zero- and few-shot prompting settings, and a systematic analysis of recurrent translation errors across selected models.

    \item[-] An extensive evaluation of 19 open- and closed-weight LLMs and MT models, examining the roles of contextual information, prompting strategy, and model size in Moroccan Darija figurative translation.

    \item[-] A layer-wise probing analysis examining how figurative meaning is represented across model layers and whether these representations distinguish figuratively appropriate meanings from literal translations.
\end{itemize}

\noindent The remainder of this paper is organized as follows: Section~\ref{related_work} reviews prior research on figurative language processing, resources, and LLM-based translation evaluation. Section~\ref{dataset} describes the construction and characteristics of MDFC. Section~\ref{methodology} presents the evaluated models, prompting settings, and evaluation metrics. Section~\ref{results} presents and discusses the experimental results. Finally, Section~\ref{conclusion} summarizes the study and outlines directions for future research.

\section{Related Work}
\label{related_work}
%This section reviews existing work on figurative language by first presenting resources and models developed for high-resource languages, followed by studies focusing on Arabic dialects.
\subsection{Figurative Language Resources and Models}
Figurative language is closely tied to culture and context, making it challenging for models to accurately interpret and translate~\citep{jhamtani-etal-2021-investigating}. Its meaning is often non-compositional and depends on linguistic and cultural contexts~\citep{chakrabarty-etal-2022-rocket}, making literal interpretation or translation insufficient to preserve the intended meaning. Consequently, many researchers have proposed dedicated datasets and evaluation frameworks.

For instance, Zhu et al. \cite{zhu-etal-2022-configure} introduced a human-annotated Chinese dataset with 12 figures of speech. They evaluated extraction, classification, and task recognition with RoBERTa-based models. The results showed that extraction and recognition tasks are still challenging. %, suggesting the need for models that localize and detect figurative language accurately.
In~\citep{chakrabarty-etal-2022-flute}, the authors introduce FLUTE, a novel dataset for figurative language understanding, including around 9,000 examples created using a model-in-the-loop framework. This dataset improved the performance of a fine-tuned T5 model compared to training on the e-SNLI dataset~\citep{camburu2018esnlinaturallanguageinference}.
Furthermore, Prystawski et al. \cite{prystawski2022psychologically} approached the metaphorical expressions using the Katz dataset~\citep{Katz01121988} by pairing each one with multiple paraphrases. They evaluated GPT-3 models using chain-of-thought prompts and showed that these reasoning prompts improve metaphor understanding. Jang et al. \cite{jang-etal-2023-figurative} tackle figurative language classification using the FLUTE dataset. They fine-tuned transformer-based models and compared them with traditional models such as logistic regression and random forest. The results showed that transformer models achieve higher F1 scores, and feature analysis indicates that this is associated with their greater focus on content words, whereas traditional models tend to be attracted by high-frequency function words, which are of little help in distinguishing between different types of figurative language. In another contribution, Kabra et al.\cite{kabra-etal-2023-multi} proposed a figurative language dataset for seven culturally diverse languages. They evaluated it on XLM-R and mBERT with fine-tuning, and GPT-3 and BLOOM with prompting. The results showed that figurative expressions are mainly culture-specific, as evidenced by their dataset analysis of culturally grounded concepts. Additionally, they experimented with few shot prompting, leading to a slight improvement, while zero-shot performance remained relatively poor.
Similarly, Bollepally et al.~\cite{bollepally2026llmsinterpretfigurativelanguage} assessed LLM's capability in interpreting figurative language. Humans and four LLMs rated sentences covering idioms, humor, sarcasm, emotions, and Gen Z slang, using 40 interpretive questions. Surface-level correlation and Representational Similarity Analysis were then used to compare human and model judgments. There was poor correspondence between humans' and machines' interpretations when dealing with context-sensitive linguistic constructions like idioms and slang. Moreover, Hosseini et al.~\cite{hosseini-etal-2026-ffe} introduced FFE-HALLU, which consists of 600 instances as a benchmark task in Persian language to evaluate idiomatic and proverbial hallucinations of language models. They assessed six multilingual LLMs on expression generation, fabricated-expression detection, and English-to-Persian translation. The results showed that most models struggled to distinguish authentic expressions from plausible fabrications and frequently hallucinated during translation. Furthermore, Ye et al.~\cite{ye-etal-2026-probing} evaluated seven LLMs using the Fig-QA~\citep{liu-etal-2022-testing} and MUNCH~\citep{tong-etal-2024-metaphor} datasets. In their methodology, they employed geometric probes for quantifying semantic alignments, used both context-dependent and context-free substitution tasks for assessing dependence on lexicons, and controlled syntactic perturbations to assess metaphor detection. The results revealed semantic shift, persistent lexical associations, and sensitivity to syntactic irregularities.%%%%%%%

Overall, research on figurative language processing has combined the construction of human-annotated and model-in-the-loop resources~\citep{zhu-etal-2022-configure,kabra-etal-2023-multi,chakrabarty-etal-2022-flute} with increasingly diverse modeling and evaluation methods, including fine-tuning~\citep{jang-etal-2023-figurative}, prompting~\citep{prystawski2022psychologically,kabra-etal-2023-multi}, human–LLM judgment comparisons~\citep{bollepally2026llmsinterpretfigurativelanguage}, and representation-based probing~\citep{ye-etal-2026-probing}. Across these methodological settings, the findings consistently indicate that current LLMs, particularly open-weight models, remain limited in their ability to capture context-sensitive and culturally grounded meanings~\citep{kabra-etal-2023-multi,bollepally2026llmsinterpretfigurativelanguage,hosseini-etal-2026-ffe}. Further systematic investigation is therefore required to determine how effectively these models process figurative expressions in low-resource and culturally specific linguistic contexts.

\subsection{Figurative Language Resources for Arabic Dialects}

\begin{table*}[!ht]
\centering
\small
\setlength{\tabcolsep}{4.5pt}
\begin{tabular}{p{3cm} l p{3cm} p{2cm} c c c}
\toprule
\textbf{Dataset} &
\textbf{Language} &
\textbf{Figurative Type} &
\textbf{Size} &
\textbf{Ctx.} &
\textbf{Translation} &
\textbf{LLM Bench.} \\
\midrule

AIPSeLEX~\cite{S_Ibrahim_2015}
& MSA, Egyptian
& Idioms, Proverbs
& 3,632
& \ding{55}
& \ding{55}
& \ding{55} \\

AFSA~\cite{elmitwally2020classification}
& MSA, Saudi
& Hyperbole, Simile, Sarcasm, Metaphor
& 2,000 (1,000 fig.; 1,000 lit.)
& \ding{51}
& \ding{55}
& \ding{55} \\

Jawaher~\citep{magdy-etal-2025-jawaher}
& 20 Arabic varieties
& Proverbs
& 10,037
& \ding{55}
& \ding{51}
& \ding{51} \\

Fann or Flop~\citep{al-ghallabi-etal-2025-fann}
& Arabic
& Poetry
& 6,984
& \ding{51}
& \ding{55}
& \ding{51} \\

ALPS~\citep{alolimat2026alpsdiagnosticchallengeset}
& Arabic
& Semantic and pragmatic phenomena
& 531
& \ding{55}
& \ding{55}
& \ding{51} \\

Yemeni Proverbs~\citep{Thmer2026}
& Yemeni Arabic, MSA
& Proverbs
& 5,252
& \ding{55}
& \ding{55}
& \ding{51} \\

Arabic Idioms~\citep{HIMDI202437}
& MSA, Arabic dialects
& Idioms
& 3,320 (1,660 fig.; 1,660 lit.)
& \ding{55}
& \ding{55}
& \ding{55} \\

Kinayat~\citep{attia-etal-2026-beyond}
& Egyptian Arabic
& Idioms
& 325
& \ding{51}
& \ding{55}
& \ding{51} \\

\hline

\textbf{MDFC (Ours)}
& \textbf{Moroccan Darija}
& \textbf{Idioms, Metaphors, Proverbs}
& \textbf{1,801}
& \ding{51}
& \ding{51}
& \ding{51} \\

\bottomrule
\end{tabular}
\caption{Comparison of Arabic figurative language datasets and benchmarks.
Ctx. and LLM Bench. indicate the inclusion of context and LLM-based evaluation,
respectively. Fig. and lit. denote figurative and literal instances, respectively.}
\label{tab:dataset_comparison}
\end{table*}
Existing Arabic dialects remain underexplored for figurative expressions, despite the availability of a few datasets~\citep{bouamor-etal-2018-madar, meftouh-etal-2015-machine, mozannar-etal-2019-neural, sajjad-etal-2020-arabench, antoun-etal-2020-arabert, khalifa-etal-2018-morphologically, ZERROUKI2017147}, with only a small number of studies exploring this direction. Table~\ref{tab:dataset_comparison} shows a comparison of Arabic figurative language benchmarks.

S.Ibrahim et al. \citep{S_Ibrahim_2015} present AIPSeLEX, a manually annotated resource containing 3,632 idioms and proverbs in Modern Standard Arabic and Egyptian Arabic for sentiment analysis. Their detection approach combines: N-gram extraction, cosine similarity, and Levenshtein distance, before assigning them polarity labels. Experiments on Arabic tweets, reviews, and comments show that the combined approach achieves 98.62\% detection accuracy. Elmitwally and Alsayat \cite{elmitwally2020classification} proposed an Arabic dataset for figurative language sentiment analysis, including a total of 2,000 texts labeled as literal and figurative, collected from Saudi dialects and Modern Standard Arabic. Three models were evaluated for this task, with Bernoulli Naive Bayes performing best, achieving a 94\% F1 score. Similarly, Magdy et al. \cite{magdy-etal-2025-jawaher} proposed a multi-dialectal dataset of 10,037 proverbs covering 20 Arabic varieties, with Arabic and English translations and explanations. Zero-shot evaluation showed that current models translate accurately but struggle with contextual and cultural explanations. Moreover, Al Ghallab et al.~\cite{al-ghallabi-etal-2025-fann} introduced the Fann or Flop dataset, a benchmark consisting of 6,984 poem–explanation pairs, evaluated on multiple open- and closed-source LLMs. The results indicated that models struggle with deep poetic and cultural understanding. Additionally, Al-Olimat and Alshareef~\cite{alolimat2026alpsdiagnosticchallengeset} present ALPS, a manually curated benchmark of 531 multiple-choice questions across 15 semantic and pragmatic tasks for evaluating Arabic linguistic reasoning. The authors evaluate 23 LLMs in a zero-shot setting, showing that Gemini-3-Flash achieves the best performance, while current models still struggle with morpho-syntactic reasoning. More recently, Thmer et al.~\cite{Thmer2026} develop a dataset that involves 5,252 Yemeni Arabic proverbs paired with MSA explanations. The authors fine-tuned AraT5v2 on the dataset and evaluated its ability to generate explanations for unseen proverbs. They also evaluated seven LLMs on 456 proverbs under zero- and few-shot settings, revealing persistent limitations in culturally grounded proverb interpretation. While Hmidi ~\cite{HIMDI202437} build a balanced dataset involving 1,660 Arabic idioms collected from online platforms and a proverb handbook, and 1,660 literal statements gathered from news, blogs, and X. Seven deep-learning and transformer-based models were evaluated, with a transformer ensemble achieving the highest accuracy of 97.8\%. Finally, in \citet{attia-etal-2026-beyond}, the authors approached the Egyptian Arabic figurative language at three levels: pragmatic use, understanding, and contextual understanding, using a dataset of 325 Egyptian Arabic idioms. 22 open- and closed-source LLMs were evaluated, and the best results were achieved by Claude 3.5 Sonnet and GPT-4o, with the highest performance on contextual understanding. This study, however, used a limited number of samples for evaluation, e.g. only 150 sentences for contextual understanding. 

Although existing studies have begun to address Arabic figurative language~\citep{magdy-etal-2025-jawaher,attia-etal-2026-beyond}, the ability of LLMs to understand and translate figurative expressions in Moroccan Darija remains largely unexplored. This represents an important research gap because figurative meaning is often non-compositional and strongly dependent on linguistic, contextual, and cultural knowledge~\citep{jhamtani-etal-2021-investigating,chakrabarty-etal-2022-rocket,kabra-etal-2023-multi}.
\section{Moroccan Darija Figurative Language Corpus}
\label{dataset}
This section describes the design and construction of MDFC, including its expression-level and context-level datasets, evaluation tasks, and annotation process.
\subsection{Dataset Design}
The MDFC dataset is designed to evaluate language and machine translation models on figurative language interpretation and translation. Specifically, it is used to assess whether they can actually distinguish the intended meaning from its literal interpretation and generate an appropriate English translation. The corpus is composed of Moroccan Darija idioms, metaphors, and proverbs, along with their English translations.

MDFC is organized into two complementary datasets: an expression-level dataset and a context-level dataset. The expression-level set presents the figurative expressions in isolation, allowing evaluation of models' capability on whether they can recover their intended meanings without the support of a sentence context.

The context-level dataset embeds figurative expressions within sentences. This dataset has been created to assess how linguistic and pragmatic context influences figurative meaning comprehension. This two-level design is proposed to support additional analysis of context-based and context-free figurative-language understanding.

Each dataset instance follows a triplet structure:
\begin{equation}
(x_i, y_i^{lit}, y_i^{fig}),
\end{equation}
where $x_i$ represents the Moroccan Darija source sentence, either from expression- or context-level sets; $y_i^{lit}$ is its word-for-word English translation, and $y_i^{fig}$ is an English translation that conveys the intended figurative meaning. The literal translation functions as an alternative distractor, whereas the figurative translation is the desired target translation.
%%%%%%%%%%%%%%%
Based on this structure, MDFC has two complementary evaluation tasks: a translation task and an MCQ task. For the first one, each model receives $x_{i}$ as input and is asked to generate the intended figurative English translation.

\noindent For the MCQ task, each original triplet is transformed to represent the following structure:
\begin{equation}
(x_i, c_i^{A}, c_i^{B}, a_i),
\end{equation}
where $c_i^{A}$ and $c_i^{B}$ denote the two candidate English translations, and $ a_i \in {A,B}$ indicates the correct answer. The two candidates correspond to a random assignment of the literal and figurative translations.
Accordingly, $(a_i=A)$ when $(c_i^{A}=y_i^{fig})$, and $(a_i=B)$ when $(c_i^{B}=y_i^{fig})$. Randomizing the positions of the two translations prevents the correct answer from being systematically associated with a particular option. Unlike the translation task, the MCQ evaluates whether a model can recognize the intended figurative meaning from the literal one without requiring models to generate the translation.

Together, the two dataset levels and evaluation tasks enable a comparative analysis of context-free and context-based figurative-language understanding, as well as the distinction between recognizing an intended translation and generating an appropriate one. 

\subsection{Expression-Level Dataset Constuction}
To construct the expression-level dataset, we first prompted GPT-5.0 and Gemini-2.0-Flash to generate Moroccan Darija idiomatic expressions. The generated candidates were manually reviewed by the annotators to remove duplicates, incorrect or unclear expressions, and examples that did not convey a figurative meaning. Valid instances were corrected and reformulated when necessary.

To increase the size and diversity of the dataset, the model-generated expressions were augmented with examples provided by the human annotators and proverbs obtained from the Moroccan Darija subset of the PALM dataset~\cite{alwajih-etal-2025-palm}. The PALM examples account for 43.22\% of the final expression-level dataset. All examples, regardless of the source, were carefully reviewed to retain only high-quality instances. The final dataset contains 423 expressions, each one paired with a literal English and intended figurative translation.
\subsection{Context-Level Dataset Construction}
The context-level dataset construction was based on the expression level, where each instance is embedded within natural contexts. Using Gemini-2.0-Flash\footnote{For contextual sentence generation, we retained only Gemini-2.0-Flash to avoid redundancy and overlapping contextual sentences that could arise from using both Gemini-2.0-Flash and GPT-5.0. Gemini-2.0-Flash was also selected for this stage because it demonstrated stronger performance on figurative language in a preliminary experiment, as manually assessed by a native speaker.}, we generate contextual sentences by incorporating the intended meanings from the Expression-level dataset into the prompt to guide the correct interpretation (Figure~\ref{prompt_template1} in Appendix). Four contextual variants are generated for each expression and manually validated by two annotators, who remove incoherent samples and correct minor errors, ensuring that all retained samples are human-verified. 

To increase linguistic diversity, particularly in terms of metaphorical expressions, we incorporate human-verified samples from the Casablanca dataset~\cite{talafha-etal-2024-casablanca} and Flores Darija~\cite{costa2022no}, from which we select only sentences containing figurative language. These resources contribute 14.60\% and 6.94\% of the final Context-level dataset, respectively. Gemini-2.0-Flash is then used to generate both literal and intended translations, which are subsequently reviewed, corrected, and reformulated by annotators. Since those expressions often do not admit literal English equivalents, annotators paraphrase the intended translations to accurately convey the underlying meaning of the source sentence. The final Context-level dataset contains 1,801 triplets, each consisting of a source sentence, a literal, and an intended English translation. 

Examples from both datasets are shown in Table~\ref{data_examples}, while their corresponding MCQ transformations are presented in Table~\ref{data_mcq_examples}.

\begin{table*}
\centering
\small
\begin{tabular}{l p{0.27\linewidth} p{0.27\linewidth} p{0.27\linewidth}}
\toprule
\textbf{Type} & \textbf{Source Sentence} & \textbf{Literal Translation} & \textbf{Figurative Translation} \\
\midrule
\multirow{2}{*}{\textbf{Expression}} 
& \foreignlanguage{arabic}{فكرشو العجينة} & There is dough in his belly. & He is an evasive person. \\ 
\cline{2-4}
& \foreignlanguage{arabic}{ماكاين غير جري عليا نجري عليك} & There is nothing but `run at me; I run at you'.  & I am quite busy these days. \\
\midrule
\multirow{2}{*}{\textbf{Context}} 
& \foreignlanguage{arabic}{هاداك باين ليا ف كرشو العجينة ماكايبانش ليا نيتو صافية.} & That one, it appears to me there is dough in his belly; his intention doesn't appear clear to me. &  I can tell that guy is hiding something; he doesn't seem like he has good intentions at all.\\
\cline{2-4}
& \foreignlanguage{arabic}{معرفتش واش نقدر نجي الصراحة، ماكاين غير جري عليا نجري عليك.} & I don't know if I can come honestly; there is nothing but running on me, I run on you. & Honestly, I am not sure I can make it. Things have been nonstop lately, and I’m busy\\
\bottomrule
\end{tabular}
\caption{Representative examples from the Expression-level and Context-level datasets.}
\label{data_examples}
\end{table*}

\begin{table*}
\centering
\small
\begin{tabular}{l p{0.23\linewidth} p{0.23\linewidth} p{0.23\linewidth} p{0.1\linewidth} }
\toprule
\textbf{Type} & \textbf{Question} & \textbf{Option A} & \textbf{Option B} & \textbf{Answer}\\
\midrule
\multirow{2}{*}{\textbf{Expression}} 
& \foreignlanguage{arabic}{فكرشو العجينة} & He is an evasive person. & There is dough in his belly. & A \\ 
\cline{2-5}
& \foreignlanguage{arabic}{ماكاين غير جري عليا نجري عليك} & I am quite busy these days. & There is nothing but `run at me; I run at you'.  &  A \\
\midrule
\multirow{2}{*}{\textbf{Context}} 
& \foreignlanguage{arabic}{هاداك باين ليا ف كرشو العجينة ماكايبانش ليا نيتو صافية.} & That one, it appears to me there is dough in his belly; his intention doesn't appear clear to me. &  I can tell that guy is hiding something; he doesn't seem like he has good intentions at all. & B\\
\cline{2-5}
& \foreignlanguage{arabic}{معرفتش واش نقدر نجي الصراحة، ماكاين غير جري عليا نجري عليك.} & Honestly, I am not sure I can make it. Things have been nonstop lately, and I’m busy & I don't know if I can come honestly; there is nothing but running on me, I run on you. &  A\\
\bottomrule
\end{tabular}
\caption{Representative MCQ examples from the Expression-level and Context-level datasets.}
\label{data_mcq_examples}
\end{table*}

\subsection{Annotation Process}
The annotation was carried out by two native speakers of Moroccan dialect (Darija), who are fluent in English. One annotator is a PhD student in computational linguistics and co-author of this work. The second annotator holds a Master's degree in Arabic–English–French translation and was recruited and compensated hourly for this task. The validation, editing, and translation of both datasets were carried out by the two annotators, who each annotated an equal portion of the dataset. Throughout the annotation process, ambiguous cases or instances requiring further verification were discussed jointly by the annotators until a consensus was reached, and the resulting annotation was retained in the final dataset.
\section{Methods and Experiments}
\label{methodology}
This section presents the experimental setup, including the evaluated models, zero-shot and six-shot prompting settings, evaluation methods for the translation and MCQ tasks, and implementation details.
\subsection{Evaluated Models}
The experiments covered 19 models selected to provide an evaluation of Moroccan Darija figurative language translation. The selected models comprise Arabic-centric, multilingual MT, and general-purpose models for both open- and closed-weights families, with parameter sizes ranging from 590M to 27B. The evaluated open-weights models include NLLB~\citep{costa2022no} (600M, 1B, and 3.3B parameters); Atlas-Chat~\citep{shang2024atlas} (2B, 9B, and 27B); AceGPT~\citep{huang-etal-2024-acegpt} (7B and 13B, in both base and chat versions); the JAIS family~\citep{sengupta2023jaisjaischatarabiccentricfoundation} (590M, 1.3B, 2.7B, and 6.7B); as well as Fanar 1.9B~\citep{team2025fanar} and ALLAM 7B~\citep{bari2025allam}. We assessed closed-weighted models as well, including GPT-4.1~\citep{openai2023gpt} and Claude~\citep{anthropic2025claude4} (both opus-4.6 and sonnet-4.6 versions).

All systems were evaluated on both the expression-level and context-level datasets under zero-shot and six-shot prompting settings. In the six-shot setting, the same six randomly selected demonstrations were provided to all evaluated systems as in-context examples. These examples, along with any other sentences containing the same figurative expressions, were removed from the corresponding evaluation datasets to prevent data leakage. NLLB models weren't included in the MCQ evaluation task given that they are designed specifically for machine translation and do not support instruction-based multiple-choice evaluation, same for few-shot experiments.

\subsection{Evaluation Metrics}

\noindent The quantitative evaluation of the MT results, combined the following four methods:

\begin{itemize}
\item[-] \textbf{Lexical similarity:} We used BLEU~\citep{papineni-etal-2002-bleu} to measure lexical overlap between the generated and reference translations. Scores were computed using SacreBLEU v2.6.0~\citep{post-2018-call}.

\item[-] \textbf{Semantic similarity:} BERTScore~\citep{zhang2019bertscore} and COMET~\citep{rei2020comet} were used to assess translation quality beyond lexical overlap. BERTScore F1 values~\citep{attia-etal-2026-beyond} were computed using the \texttt{microsoft/deberta-xlarge-mnli} model and the \texttt{bert-score} library v0.3.13. COMET scores~\citep{comet,rei2020comet} were computed using the \texttt{McGill-NLP/ssa-comet-mtl} model and the \texttt{unbabel-comet} library v2.2.7.

\item[-] \textbf{LLM-as-a-judge:} Translation quality was assessed along two dimensions: adequacy, and fluency~\citep{zheng2023judging}.\footnote{Adequacy corresponds to meaning preservation and fluency to linguistic quality; detailed definitions and rating scales are provided in Table~\ref{criteria_llm_ev}.} Each dimension was rated on a five-point Likert scale. Gemini-3.8-flash model was then used as a judge to evaluate the translations.

\item[-] \textbf{Human evaluation:} A subset of the generated translations was evaluated by human annotators to assess the reliability of the LLM-as-a-judge setup.
\end{itemize}
\begin{table*}[!htbp]
\centering
\small
\begin{tabular}{p{0.07\linewidth} p{0.23\linewidth} p{0.6\linewidth}}
\toprule
\textbf{Measure} & \textbf{Description} & \textbf{Scale (1--5)} \\
\midrule

\textbf{Adequacy} &
Measures how the translation preserves semantic correctness. &
\begin{minipage}[t]{\linewidth}
5: Perfectly conveys the intended meaning.\\
4: Mostly accurate; minor nuances missed without altering the overall meaning.\\
3: Conveys general meaning, but some important details are missing \\
2: Mostly incorrect translation\\
1: Completely incorrect or misleading translation \\
\end{minipage} \\

\midrule

\textbf{Fluency} &
Measures the grammaticality and fluency of the English translation. &
\begin{minipage}[t]{\linewidth}
5: Fully natural, native-like English\\
4: Slightly unnatural but still fluent\\
3: Understandable but somewhat awkward\\
2: Clearly unnatural or poorly phrased\\
1: Very unnatural or broken English\\
\end{minipage} \\
\bottomrule
\end{tabular}
\caption{Evaluation criteria used for LLM-as-a-judge scoring.}
\label{criteria_llm_ev}
\end{table*}
For the MCQ task, we considered accuracy as the model performance measure, defined as the percentage of cases in which the model selected the intended figurative translation.
\subsection{Implementation details}

\noindent All open-weight models were evaluated using deterministic greedy decoding. The maximum generation length was set to 256 tokens for the translation and to 16 tokens for the MCQ task, in which models were instructed to generate only the selected answer option (A or B). Inference was performed in BF16 precision on a single NVIDIA H100 GPU with 80 GB of HBM3 memory~\citep{Kissami2025}.

\noindent The closed-weights models were accessed through their official APIs. GPT-4.1 was evaluated using the OpenAI API, whereas Claude Opus 4.6 and Claude Sonnet 4.6 were evaluated using the Anthropic API in July 2026.

\begin{table*}[!ht]
\centering
\small
\begin{tabular}{p{2.8cm}ccccccccccc}
\toprule
\textbf{Model}
& \multicolumn{5}{c}{\textbf{Expression-Level}}
& \multicolumn{5}{c}{\textbf{Context-Level}} \\
\cmidrule(lr){2-6}\cmidrule(lr){7-11}
& \textbf{BLEU}
& \textbf{BERT}
& \textbf{COM.}
& \multicolumn{2}{c}{\textbf{LLM-as-a-judge}}
& \textbf{BLEU}
& \textbf{BERT}
& \textbf{COM.}
& \multicolumn{2}{c}{\textbf{LLM-as-a-judge}} \\%chrf++
\cmidrule(lr){5-6}\cmidrule(lr){10-11}
&
&
&
&
Adeq.
& Flu.
&
&
&
&
Adeq.
& Flu.

\\
\midrule
NLLB 600M 
& 0.57 & 54.44 & 34.34 & 1.35 & 2.93 
& 7.26 & 62.44 & 37.77 & 1.37 & 2.52 
\\

NLLB 1.3B 
& 0.86 & 55.58 & 35.48 & 1.40 & 3.21 
& 8.44 & 64.33 & 39.81 & 1.55 & 2.88 
\\

NLLB 3.3B 
& 0.72 & 56.53 & 36.41 & 1.45 & 3.29  
& 8.87 & 65.36 & 40.83 & 1.62 & 3.15 
\\

\hline
Atlas-Chat 2B 
& 0.88 & 55.90 & 37.44 & 1.67 & 3.44 
& 7.88 & 59.67 & 42.72 & 1.86 & 2.81 
\\

Atlas-Chat 9B 
& 2.22 & 59.12 & 38.99 & 2.25 & 3.85 
& 14.55 & 71.86 & 45.88 & 2.76 & 3.80 
\\

Atlas-Chat 27B 
& 1.55 & 58.40 & 39.30 & 2.37 & 3.76  
& 13.51 & 71.00 & 45.48 & 2.87 & 3.74 
\\

\hline
ALLaM-7B
& 0.85 & 49.73 & 34.53 & 1.85 & 2.84  
& 12.77 & 67.81 & 43.89 & 2.55 & 3.20
\\

Fanar 9B
& 0.83 & 58.26 & 36.81 & 1.57 & 3.03  
& 6.93 & 63.25 & 42.59 & 1.69 & 2.22 
\\

\hline
AceGPT-7B 
& 0.58 & 54.97 & 39.05 & 1.64 & 3.22 
& 7.21 & 63.08 & 41.09 & 1.74 & 3.25  
\\

AceGPT-13B 
& 0.91 & 54.53 & 37.67 & 1.65 & 3.38  
& 6.70 & 61.63 & 40.98 & 1.66 & 3.01  
\\

AceGPT-7B-chat 
& 1.43 & 57.93 & 37.84 & 1.48 & 3.34 
& 6.02 & 61.89 & 41.40 & 1.48 & 2.92  
\\

AceGPT-13B-chat 
& 1.59 & 56.80 & 37.32 & 1.49 & 3.14  
& 7.63 & 63.42 & 41.73 & 1.58 & 2.71 
\\

\hline
jais-family-590m 
& 0.17 & 52.54 & 34.90 & 1.19 & 2.52 
& 2.70 & 57.62 & 36.83 & 1.23 & 2.43  
\\

jais-family-1p3b 
& 0.32 & 52.77 & 35.27 & 1.29 & 2.46 
& 4.46 & 60.25 & 39.03 & 1.44 & 2.26
\\

jais-family-2p7b 
& 0.43 & 55.42 & 36.91 & 1.54 & 2.54 
& 6.49 & 64.62 & 42.38 & 1.81 & 2.53 
\\

jais-family-6p7b 
& 0.59 & 57.63 & 38.36 & 1.84 & 2.90 
& 8.40 & 67.56 & 43.94 & 2.16 & 2.70 
\\
\hline

GPT-4.1
& \textbf{5.53} & \textbf{67.23} & \textbf{43.26} & \textbf{3.13} & \textbf{4.05} 
& \textbf{21.83} & \textbf{77.58} & \textbf{51.13} & \textbf{3.82} & \textbf{4.14} 
\\

Claude-opus-4.6
& 4.40 & 64.00 & 41.82 & 2.96 & 3.75
& 21.75 & 76.38 & 49.89 & 3.68 & 4.02 
\\

Claude-sonnet-4.6
& 2.89 & 60.54 & 41.12 & 2.67 & 3.34 
& 16.86 & 73.40 & 48.76 & 3.18 & 3.42 
\\

\bottomrule
\end{tabular}
\caption{Evaluation of models on figurative translation. BLEU, BERTScore (BERT), and COMET (COM.) are reported on a 0--100 scale. LLM-as-a-judge evaluates translations using Adequacy (Adeq.) and Fluency (Flu.), each on a 1--5 scale.}
\label{tab1}
\end{table*}

\begin{table*}[!t]
\centering
\small
%\large
\begin{tabular}{lccc} 
\toprule
\textbf{Model} & \textbf{Expression-Level} & \textbf{Context-Level} & \textbf{$\Delta$ Accuracy} \\
\midrule

Atlas-Chat 2B & 40.66 & 55.75 & +15.09\\
Atlas-Chat 9B & 39.24 & 52.47 & +13.23\\
Atlas-Chat 27B & 37.59 & 51.36 & +13.77\\
\hline
ALLaM-7B & 31.91 & 46.53 & +14.62 \\
Fanar & 38.06 & 55.75 & +17.68\\
\hline
AceGPT-7B & 50.12 & 47.64 & -2.48\\
AceGPT-13B & 46.57 & 52.92 & +6.35\\
AceGPT-7B-chat & 25.53 & 41.87 & +16.34\\
AceGPT-13B-chat & 47.99 & 51.75 & +3.76\\
\hline
jais-family-590m & 40.90 & 43.36 & +2.46\\
jais-family-1p3b & 49.41 & 42.48 & -6.93\\
jais-family-2p7b & 47.99 & 51.19 & +3.20\\
jais-family-6p7b & 3.78 & 14.88 & +11.10\\
\hline
%GPT-5.1 & \textbf{70.56} & \textbf{81.84} & +11.28\\
GPT-4.1 & \textbf{60.76} & 62.13 & +1.37\\
%Gemini-2.5-Flash & 44.39 & 49.80 & +5.41\\
%Gemini-2.0-Flash & 53.27 & 71.90 & +18.63\\
Claude-opus-4.6 & 44.68 & \textbf{66.51} & +21.83 \\
Claude-sonnet-4.6 & 39.48 & 57.85 &  +18.37\\
\bottomrule
\end{tabular}
\caption{Accuracy (\%) of the evaluated models on the expression-level and context-level MCQ benchmarks. $\Delta$ Accuracy denotes context-level minus expression-level accuracy, in percentage points.}
\label{accuracy}
\end{table*}

%\section{Results}
%In this section, we present the quantitative results, human evaluation, and probing analysis of the evaluated models.
\section{Results and Discussion}
\label{results}
\subsection{Zero-shot Translation Performance}
Table~\ref{tab1} shows the zero-shot results of figurative translation for the evaluated models in terms of expression-level and context-level datasets. Model GPT-4.1 obtains the highest scores in both evaluation settings. Claude-opus-4.6 follows closely, ranking second in BLEU, BERTScore, COMET, and adequacy. Among the evaluated models, the closed-weight models outperform the open-weight ones in BLEU, BERTScore, COMET, and adequacy, implying better capacity in conveying the intended figurative meaning compared to the open-weight models. However, this advantage is less consistent for fluency, where some Atlas-Chat variants outperform the Claude models. GPT-4.1 achieves adequacy scores of 3.13 and 3.82 at expression level and context level tests, respectively. Thus, indicating moderate to good performance for preserving the figurative meanings of Moroccan Darija. Nevertheless, the lower adequacy scores of open-weight models indicate that this capability varies considerably across systems, making figurative translation of Moroccan Darija remain challenging, \textbf{RQ1}.

The relationship between parameter count and translation performance varies across model families and evaluation metrics. NLLB models improve consistently on every metric except expression-level BLEU, which decreases when scaling from 1.3B to 3.3B. Jais models show consistent improvements in BLEU, BERTScore, COMET, and adequacy at both levels, while fluency initially decreases from 590M to 1.3B before improving for larger variants. However, Atlas-Chat 9B outperforms Atlas-Chat 27B on most metrics, despite the latter achieving higher adequacy. Similarly, increasing the parameter count from AceGPT-7B to AceGPT-13B reduces performance across all context-level metrics. This decline does not extend to their chat variants, where the 13B model achieves higher context-level scores on all metrics except fluency. These findings indicate that increasing parameter count does not consistently improve figurative translation performance, addressing \textbf{RQ4}.
\subsection{Effect of Context on Translation}
Performance generally improves when figurative expressions are presented within contextualized sentences. As shown in Table~\ref{tab1}, all models achieve higher BLEU, BERTScore, and COMET scores on the context-level dataset. Adequacy also improves for all models except AceGPT-7B-chat, whose score remains unchanged, with particularly notable gains for Claude-opus-4.6, ALLaM-7B, and GPT-4.1, whose scores increase by 0.72, 0.70, and 0.69 points, respectively. A similar overall trend is observed in the MCQ results presented in Table~\ref{accuracy}, although accuracy decreases for AceGPT-7B and jais-family-1p3b. These findings suggest that contextual information generally provides semantic and pragmatic cues that help models identify and preserve the intended meaning of Moroccan Darija figurative expressions (\textbf{RQ2}).

Nevertheless, most open-weight models register reduced fluency scores at the context level relative to those at the expression level. The exceptions are ALLaM-7B and AceGPT-7B, whose fluency scores increase from 2.84 to 3.20 and from 3.22 to 3.25, respectively. All three closed-weight models also show improved fluency with context. These results suggest that improvements in adequacy with context do not necessarily coincide with improvements in fluency, particularly among open-weight models.\\

\subsection{Figurative-Translation Selection Results}
The accuracy of figurative English translation detection using multiple-choice questions is presented in Table~\ref{accuracy}. On the expression-level dataset, the highest accuracy was achieved by GPT-4.1 (60.76\%), followed by AceGPT-7B and jais-family-1p3b. Meanwhile, at the context level, the best-performing model was Claude-opus-4.6 (66.51\%), then GPT-4.1 and Claude-sonnet-4.6. Closed-weight models exhibit superior predictive abilities. Nonetheless, some open-weight ones perform competitively, especially on the expression-level dataset.

Contextual information improves MCQ accuracy for most models. Both Claude variants obtain remarkable gains from the use of context, with the largest difference improvements compared to the others. However, AceGPT-7B and jais-family-1p3b exhibit negative $\Delta$ values, indicating that context does not benefit all models equally. Some models also achieve moderate MCQ accuracy despite their relatively low translation adequacy scores. For example, Fanar achieves 55.74\% MCQ accuracy on the context-level dataset despite receiving an adequacy score of only 1.69. Similarly, jais-family-1p3b obtains 49.41\% MCQ accuracy on the expression-level dataset but with an adequacy score of only 1.29. These results indicate that models may recognize the intended figurative meaning when presented with the literal translation alternative while struggling to generate an adequate translation independently. The statistical significance of differences between models is further examined using pairwise McNemar tests, as detailed in Appendix~\ref{SA}.\\

\subsection{Few-shot Translation Performance}

\label{few_shot}
Table~\ref{tab5} shows that the evaluated closed-weight models outperform open-weight models in BLEU, BERTScore, COMET, and adequacy at both the expression and context levels under few-shot prompting. Claude-opus-4.6 achieves the highest scores across all expression-level metrics. At the context level, it leads in BLEU and BERTScore, while GPT-4.1 achieves the highest COMET, adequacy, and fluency scores. Among open-weight models, Atlas-Chat 9B achieves the highest expression-level BLEU, COMET, and fluency scores, while ALLaM-7B achieves the highest expression-level BERTScore and adequacy, and the strongest context-level results across all metrics except fluency.

Compared with zero-shot prompting (Table~\ref{tab1}), the benefits of in-context examples vary across models and metrics. ALLaM-7B and AceGPT-7B-chat improve across all translation metrics at both levels. All three closed-weight models also achieve higher adequacy and fluency scores at both levels, although their automatic metric scores do not uniformly increase. Claude-sonnet-4.6 shows substantial improvements: its adequacy increases from 2.67 to 3.39 at the expression level and from 3.18 to 3.89 at the context level, while fluency rises from 3.34 to 4.91 and from 3.42 to 4.81, respectively.

Several open-weight models exhibit mixed results or declines. Fanar improves in expression-level BLEU but decreases on all other expression-level metrics. AceGPT-13B and AceGPT-13B-chat also decline across all expression-level metrics. Atlas-Chat 27B improves on most expression-level metrics but decreases across all context-level metrics. Furthermore, all Jais variants obtain lower COMET and adequacy scores for both benchmarks despite higher fluency scores. This divergence indicates that in-context examples can improve the linguistic quality of translations without improving preservation of figurative meaning. Qualitative inspection identifies repetitive or hallucinated outputs for Fanar, incomplete or overly literal translations for some AceGPT variants, semantic misinterpretations for Jais, and conversational-style generations for Atlas-Chat 27B. These observations provide further insight into the limitations of few-shot prompting for these models.

The MCQ results in Table~\ref{accuracy2} show that improvements in meaning selection do not necessarily coincide with improvements in translation generation. Claude-sonnet-4.6 achieves the highest few-shot MCQ accuracy at both levels, followed by GPT-4.1. Their accuracies increase substantially relative to zero-shot prompting, whereas Claude-opus-4.6 improves at the expression level but decreases from 66.51\% to 50.84\% at the context level. Similarly, Fanar and Atlas-Chat 27B achieve higher MCQ accuracy at both levels despite declines in parts of their translation performance. These differences suggest that selecting an interpretation and generating an adequate figurative translation place different demands on the evaluated models.

The benefits of context under few-shot prompting are also uneven. Although all models obtain higher context-level BLEU and COMET scores than expression-level scores, adequacy decreases for Atlas-Chat 27B and AceGPT-7B, and fluency decreases for several models. Context-level MCQ accuracy is also lower for five models. Thus, contextual information generally benefits some aspects of translation, but its advantages do not extend consistently across models, metrics, and tasks (\textbf{RQ2}).

Overall, the mixed results suggest that the benefits of context and few-shot prompting depend on the model and task. Some models produce more accurate figurative translations when given context or examples, whereas others show little improvement or generate less accurate translations. Differences in training objectives, training data, and adaptation to instruction following or conversational tasks may help explain these patterns. For example, Claude-opus-4.6’s reasoning capabilities may help it interpret contextual information, contributing to its higher translation adequacy with context and few-shot prompting. However, its context-level MCQ accuracy decreases under few-shot prompting compared with zero-shot prompting. These contrasting results suggest that each model’s training and capabilities may influence whether additional context and examples improve performance on a particular task.
\begin{table*}[!t]
\centering
\small
\begin{tabular}{p{2.8cm}cccccccccc}
\toprule
\textbf{Model}
& \multicolumn{5}{c}{\textbf{Expression-Level}}
& \multicolumn{5}{c}{\textbf{Context-Level}} \\
\cmidrule(lr){2-6}\cmidrule(lr){7-11}

& \textbf{BLEU}
& \textbf{BERT}
& \textbf{COM.}
& \multicolumn{2}{c}{\textbf{LLM-as-a-judge}}
& \textbf{BLEU}
& \textbf{BERT}
& \textbf{COM.}
& \multicolumn{2}{c}{\textbf{LLM-as-a-judge}} \\
\cmidrule(lr){5-6}\cmidrule(lr){10-11}
&
&
&
&
\textbf{Adeq.}
& \textbf{Flu.}
%& \textbf{Ctx.}
&
&
&
&
\textbf{Adeq.}
& \textbf{Flu.}
\\
\midrule
Atlas-Chat 2B 
& 2.07 & 60.26 & 37.12 & 1.75 & 3.86 
& 9.94 & 68.50 & 43.56 & 2.14 & 3.58
\\

Atlas-Chat 9B 
& 4.26 & 62.93 & 38.28 & 2.37 & 4.35 
& 13.19 & 71.57 & 45.96 & 2.88 & 4.09
\\

Atlas-Chat 27B 
& 2.90 & 60.69 & 37.06 & 2.40 & 3.96  
& 4.59 & 44.93 & 38.13 & 1.77 & 1.99
\\

\hline
ALLaM-7B
& 3.58 & 63.73 & 39.87 & 2.45 & 3.93  
& 16.08 & 73.66 & 47.77 & 3.08 & 4.11 
\\

Fanar 
& 1.68 & 41.19 & 16.97 & 1.31 & 1.56 
& 5.46 & 60.03 & 36.68 & 2.21 & 3.34 
\\

\hline
AceGPT-7B 
& 2.35 & 58.21 & 39.21 & 1.98 & 4.26
& 6.38 & 63.33 & 39.10 & 1.89 & 4.30
\\

AceGPT-13B 
& 0.13 & 33.25 & 21.89 & 1.11 & 1.19
& 6.35 & 61.62 & 38.48 & 1.85 & 4.14
\\

AceGPT-7B-chat 
& 2.54 & 60.54 & 38.62 & 1.67 & 4.07 
& 7.49 & 64.28 & 43.03 & 1.67 & 3.62 
\\

AceGPT-13B-chat 
& 1.41 & 47.44 & 24.10 & 1.39 & 2.22 
& 8.57 & 65.47 & 43.44 & 1.83 & 3.61
\\

\hline
jais-family-590m 
& 0.88 & 55.23 & 30.76 & 1.13 & 3.65 
& 2.56 & 58.54 & 34.03 & 1.18 & 3.23
\\

jais-family-1p3b 
& 0.67 & 56.44 & 32.31 & 1.24 & 3.55 
& 3.07 & 59.88 & 35.52 & 1.36 & 3.34
\\

jais-family-2p7b 
& 1.26 & 58.73 & 33.76 & 1.48 & 3.79
& 5.54 & 64.01 & 39.32 & 1.69 & 3.47
\\

jais-family-6p7b 
& 0.52 & 58.68 & 34.94 & 1.75 & 3.95 
& 6.35 & 65.54 & 40.72 & 2.00 & 3.63
\\
\hline

GPT-4.1
& 7.82 & 70.12 & 42.48 & 3.75 & 4.89 
& 18.64 & 78.02 & \textbf{51.94} & \textbf{4.28} & \textbf{4.95} 
\\

Claude-opus-4.6
& \textbf{9.74} & \textbf{70.52} & \textbf{43.10} & \textbf{3.84} & \textbf{4.93}
& \textbf{22.66} & \textbf{78.24} & 50.71 & 4.23 & 4.80 
\\

Claude-sonnet-4.6
& 6.86 & 67.97 & 40.79 & 3.39 & 4.91 
& 20.29 & 76.70 & 49.66 & 3.89 & 4.81 
\\
\bottomrule
\end{tabular}
\caption{Evaluation of models on figurative translation in few-shot settings. BLEU, BERTScore (BERT), and COMET are reported on a 0--100 scale. LLM-as-a-judge evaluates translations using Adequacy (Adeq.) and Fluency (Flu.), each on a 1--5 scale.}
\label{tab5}
\end{table*}

\begin{table*}[t]
\centering
\small
\begin{tabular}{lccc}
\toprule
\textbf{Model} & \textbf{Expression-Level} & \textbf{Context-Level} & \textbf{$\Delta$ Accuracy} \\
\midrule

Atlas-Chat 2B & 41.73 & 46.86 & +5.13\\
Atlas-Chat 9B & 73.86 & 71.08 & -2.78\\
Atlas-Chat 27B & 83.93 & 73.99 & -9.94 \\
\hline
ALLaM-7B & 70.02 & 63.73 & -6.29 \\
Fanar & 77.22 & 88.51 & +11.29 \\
\hline
AceGPT-7B & 50.12 & 52.13 & +2.01\\
AceGPT-13B & 50.12 & 52.13 & +2.01\\
AceGPT-7B-chat & 50.36 & 62.89 & +12.53\\
AceGPT-13B-chat & 62.80 & 67.54 & +4.74\\
\hline
jais-family-590m & 50.12 & 51.96 & +1.84\\
jais-family-1p3b & 50.36 & 52.47 & +2.11\\
jais-family-2p7b & 50.36 & 50.22 & -0.14\\
jais-family-6p7b & 50.12 & 49.27 & -0.85\\
\hline
%GPT-5.1 & 96.44 & \textbf{98.10} & +1.66\\
GPT-4.1 & 95.92 & 97.98 & +2.06\\
%Gemini-2.5-Flash & 47.39 & 52.03 & +4.64\\
%Gemini-2.0-Flash & \textbf{97.63} &  97.27 & -0.36\\
Claude-opus-4.6 & 50.36 & 50.84 & +0.48  \\
Claude-sonnet-4.6 & \textbf{97.12} & \textbf{98.21} & +1.09  \\

\bottomrule
\end{tabular}
\caption{Accuracy (\%) of the evaluated models on the expression-level and context-level MCQ benchmarks case of few shot prompting. $\Delta$ Accuracy denotes the performance difference between both settings.}
\label{accuracy2}
\end{table*}

\subsection{Human Evaluation}
We performed a human evaluation using a 5-point Likert scale. Three native speakers of Moroccan dialect, fluent in English\footnote{The three human evaluators are currently pursuing doctoral studies in Computational Linguistics}, participated in this task. They rated the model outputs according to the guidelines shown in Figure~\ref{prompt2}, we randomly selected 20 examples from each model, resulting in a total of 380 evaluated sentences. Inter-annotator agreement among human annotators reached 0.57 as measured by Krippendorff’s $\alpha$~\citep{krippendorff2011computing}. Additionally, to assess the reliability of LLM as a judge, we have introduced it as an evaluator, instructing Gemini-3.8-flash to follow the same evaluation guidelines as humans but provided with the gold reference. Inter-annotator agreement, including LLM as evaluator, slightly increased to a 0.62 value. \\

To further assess this alignment, we calculated the Spearman correlation between the LLM's ratings and the average scores of humans, obtaining a strong correlation of $\rho = 0.82$,. Pairwise correlations between the LLMs and each annotator range from 0.69 to 0.79, indicating a generally stable agreement across annotators. 

Table~\ref{human_ev} highlights the comparison of mean ratings of evaluated models based on human and LLM-as-a-judge. Both evaluators classify the performance of Claude-opus-4.6 and GPT-4.1 as the best. Claude-opus-4.6 receives the highest score from both humans and the LLM (3.55 and 3.80, respectively), followed by GPT-4.1 (3.52 and 3.70). However, the ranking differs for Claude-sonnet-4.6 and Atlas-Chat 27B: human annotators rank Atlas-Chat 27B higher (3.32 vs. 3.23), whereas the LLM assigns a higher score to Claude-sonnet-4.6 (3.40 vs. 3.25).

Among the open-source models, Atlas-Chat 27B receives the highest scores from both evaluators, followed by Atlas-Chat 9B and ALLaM-7B. Both evaluation methods also show consistent improvements with increasing model size within the Atlas-Chat and JAIS families. The NLLB models receive relatively low scores from both evaluators, although their ordering differs. JAIS-590M obtains the lowest score under both human and LLM evaluation (1.15 and 1.05, respectively).

The LLM assigns lower scores than human annotators to 13 of the 19 models. The largest differences occur for AceGPT-13B-chat (2.51 vs. 2.05), JAIS-2.7B (2.20 vs. 1.80), and NLLB-1.3B (1.83 vs. 1.50). Conversely, the LLM assigns higher scores to all three closed-weights models, as well as NLLB-600M, NLLB-3.3B, and JAIS-6.7B. Nevertheless, the differences in average scores remain below half a point on the 5-point scale for all models.

Overall, the statistical correlation and agreement between human and LLM judgments, together with the relatively small differences in average scores, support the suitability of the LLM-as-a-judge for comparative evaluation in this benchmark. Although the evaluators sometimes differ in their model rankings, they generally agree on the relative quality of the evaluated systems, suggesting that LLM-based evaluation can serve as a scalable approach to assessing translation quality.

\begin{table}[!t]
\centering
\begin{tabular}{lcc}
\hline
\textbf{Model} & \textbf{Human} & \textbf{LLM} \\
\hline
NLLB 600M  & 1.52 &  1.55 \\
NLLB 1.3B   & 1.83 & 1.50  \\
NLLB 3.3B  & 1.65 & 1.70  \\ \hline
Atlas-Chat 2B  & 2.67 & 2.40 \\
Atlas-Chat 9B  & 2.97 & 2.90  \\
Atlas-Chat 27B  & 3.32 & 3.25\\ \hline
ALLaM 7B  & 2.85 &  2.65 \\
Fanar 9B & 2.07 & 1.95  \\ \hline
AceGPT-7B  & 2.43 & 2.25  \\
AceGPT-13B  & 2.33 &  2.10 \\
AceGPT-7B-chat  & 1.95 &  1.75 \\
AceGPT-13B-chat  & 2.51 & 2.05  \\ \hline
jais-family-590m  & 1.15 & 1.05  \\
jais-family-1p3b  & 1.82 & 1.60  \\
jais-family-2p7b  & 2.20 & 1.80  \\
jais-family-6p7b  & 2.23 &  2.30 \\ \hline
%GPT-5.1  & \textbf{4.41} & 3.50 \\
GPT-4.1  & 3.52 & 3.70  \\
%Gemini-2.5-Flash  & 3.78 & 3.60 \\
%Gemini-2.0-Flash  & \textbf{4.58} & \textbf{4.65} \\
Claude-opus-4.6  & \textbf{3.55} & \textbf{3.80}  \\
Claude-sonnet-4.6  & 3.23  & 3.40\\
\hline
\end{tabular}
\caption{Comparison of models using human and LLM-based evaluation.}
%\vspace{-10em}
\label{human_ev}
\end{table}

\subsection{Qualitative Results}
\label{q_r}
Table~\ref{qualitative_results} presents output translations from `GPT-4.1', `Atlas-Chat 27B', and `jais-family-590m' across both datasets. The selected models reflect different performance levels observed in the quantitative evaluation, corresponding respectively to the strongest-performing model, a competitive open-source system, and a weaker small-scale model.\\
`GPT-4.1' consistently preserves the intended figurative meaning across most examples, producing the most natural-sounding translations. For instance, it accurately captures the core idiom in the second expression-level example. It demonstrates similar proficiency in the first contextual example, delivering a fluent translation that maintains the source sentence's emotional tone. However, despite its robust performance, the model occasionally leans too literal. This is evident in the final contextual example, where it mirrors the source's metaphorical structure too closely rather than adapting it into a more natural English equivalent.
`Atlas-Chat 27B' successfully captures the figurative essence of several test cases, producing clear and intelligible translations on both datasets. Notably, in the fourth expression-level example, the model correctly identifies and generates the equivalent English proverb, `What goes around comes around.' However, it occasionally clings too tightly to the source sentence's lexical structure. In the third expression-level example, for instance, the concept of "creating obstacles" is entirely lost. A similar issue occurs in the final contextual example, where the phrasing remains overly literal.
`jais-family-590m' delivers the weakest overall performance, frequently relying on word-for-word translation at the expense of figurative meaning. For example, it fails to convey the intended meaning on the third expression-level translation. In other cases, the model abandons the source semantics entirely, generating irrelevant outputs. This tendency is highly apparent in the first expression-level example and resurfaces across several contextual examples, where the model produces unstable or hallucinated text despite having the surrounding context.
\begin{table*}[!t]
\centering
\small
\setlength{\tabcolsep}{4pt}
\renewcommand{\arraystretch}{1.2}

\begin{tabular}{p{3.2cm}|p{3.2cm}|p{2.8cm}|p{2.8cm}|p{2.8cm}}
\hline
\hline
\multicolumn{5}{c}{\textbf{Expression-level}} \\
\hline
\textbf{Source Sentence} & \textbf{Gold Reference} & \textbf{GPT} & \textbf{Atlas-Chat} & \textbf{JAIS} \\
\hline
\textarabic{ما قدو فيل، زادوه فيلة} & Already struggling with a massive burden, even more weight has now been added. & He couldn't handle one elephant, and they gave him more elephants. & It wasn't enough, they added more elephants & What have you done to them? They're dead!  \\
\hline
\textarabic{جا يكحلها عماها} & He tried to fix the situation but ended up making it worse. & He tried to fix it, but made it worse. & Add insult to injury & jaklha kulthamah \\
\hline
\textarabic{كيْحطّ العصا فالرويضة} & He blocks the situation & He’s putting a stick in the wheel. (He’s causing trouble or creating obstacles.) & He puts the stick in the hole & Put your stick in your garden \\
\hline
\textarabic{الدنيا دوارة} & What goes around comes around. & Life goes around. & What goes around comes around & The world is turning \\

\hline
\hline
\multicolumn{5}{c}{\textbf{Context-level}} \\
\hline
\textbf{Source Sentence} & \textbf{Gold Reference} & \textbf{GPT} & \textbf{Atlas-Chat} & \textbf{JAIS} \\
\hline
\textarabic{ملي شفتو خدا حقو برد ليا القلب.} & Seeing him get what he deserved was satisfying. & When I saw him get what he deserved, it eased my heart. & When I saw he got his due it warmed my heart. & I saw him take his rightful place and I felt like he had taken my heart. \\
\hline
\textarabic{فهاذ الاجتماع كل واحد كيلغي بلغاه، ما غايخرج منو والو.} & In this meeting, everyone is only interested in their own point of view; nothing will come of it. & In this meeting, everyone is just talking over each other; nothing will come out of it. &  In this meeting, everyone cancels his membership, so no one will benefit. &  In this meeting, everyone will speak with him and he won't leave anything behind or any waste of time.\\
\hline
\textarabic{سولني على داك الشي؟ ماتدي مني لا حق ولا باطل.} & Did you ask me about that thing? You won't get any information or a straight answer out of me. & He asked me about that thing? You won't get anything out of me, right or wrong. & Ask me about that? You'll get no sense out of me. & What do you mean by no right and wrong, but only what I want is my way of life (I live according to the rules). \\
\hline
\textarabic{هاداك كيشري الحوت في البحر، مازال ما بنا الدار و كيهضر على فراشها.} & He's paying for something that doesn't exist yet; he still hasn't built the house, and he's talking about the furniture. & That guy buys fish while still at sea; he hasn’t even built the house yet and he’s already talking about the furniture. & He buys fish at the sea, hasn't even built a house yet and he's talking about furniture. & This one is selling fish in water, but we haven't been back home yet and he's talking about his fourniture.\\
\hline
\textarabic{مالو كيغطي على الشمس بالغربال، راه كلشي عارف.} & What's wrong with him? He's trying to hide something obvious; everyone knows. & Why is he trying to cover the sun with a sieve? Everyone knows the truth. & He pretends to cover the sun with a sieve, everyone knows. & [The sun] covers up everything with sand, everybody knows that. All of you know it! The truth is not hidden from anyone! \\
%& \textarabic{دير المهدي فكتافك.} & Rely on Mehdi & Put the pillow under your shoulder. &  &  \\
%\cline{2-6}
%& \textarabic{منين طاحت عليك المحنة؟}  &  \\
%\cline{2-6}
%& \textarabic{مالو كيغطي على الشمس بالغربال، راه كلشي عارف.} & What's wrong with him? He's trying to hide something obvious; everyone knows. & Why is he trying to cover the sun with a sieve? Everyone knows the truth. &  &  \\
%\cline{2-6}
%& \textarabic{}  &  \\

\hline
\end{tabular}
\caption{Qualitative translation examples from GPT-4.1, Atlas-Chat 27B, and jais-family-590m on the expression-level and context-level datasets.}
\label{qualitative_results}
\end{table*}

\subsection{Error Analysis}
\label{analysis_s}
To characterize recurring failure patterns \textbf{(RQ3)}, we randomly selected 100 translation examples per model and manually annotated the outputs of three representative models: GPT-4.1, Atlas-Chat-9B, and jais-family-590m. These models represent a high-performing closed-weights model, a relatively strong open-weights model, and a lower-performing open-weights model, respectively. Outputs exhibiting multiple errors were assigned more than one error category. Consequently, the percentages in Table~\ref{error_analysis} are calculated relative to the total number of category assignments for each model.

GPT-4.1 produced the greatest number of correct translations, 57 out of 100 samples labelled as correct. Literal translation is its dominant failure mode, accounting for 43 assignments (40.2\%), followed by semantic shifts with seven occurrences (6.5\%). Other types of errors, such as hallucination, omission, etc., were not observed. Although GPT-4.1 achieved better results, the model sometimes translated the context perfectly but failed to interpret the figurative expressions semantically; rather, it provided literal translations without changing the surface form.

The Atlas-Chat-9B produces 34 accurate translations of Darija to English. The model produces literal translations, which constitute the most common error type, with 41 errors, then semantic shift with 27 errors (25.2\%). These errors include lexical substitutions that alter the source meaning and misinterpretations of figurative imagery. For example, a proverb conveying “like mother, like son” is translated as a child wearing a bib to resemble a baby, losing the intended meaning of resemblance between a mother and her son. The model also occasionally returns Arabic or reproduces the Darija sentence instead of translating it into English, contributing to the 5 assignments (4.7\%) in the Other category. No hallucinations and/or omissions were identified in the model’s output.

In contrast, jais-family-590m produces only 3 correct translations. Hallucination dominates its annotation distribution, with 59 assignments (52.7\%), while literal translation and semantic shift each account for 18 assignments (16.0\%). Its outputs frequently introduce unrelated events or unsupported details. The annotation comments also reveal failures to follow the translation instruction: some outputs reformulate the prompt, respond to the source sentence, or explain the expression instead of translating it. Others contain Arabizi or mixed-language text rather than an English translation. These additional failure modes contribute to 13 assignments (11.6\%) in the Other category, alongside one omission (0.9\%).

The error patterns differ across the three models. GPT-4.1 and Atlas-Chat-9B primarily struggle to move beyond literal wording and preserve figurative meaning, whereas jais-family-590m frequently fails to generate a relevant translation. These observations highlight both the difficulty of interpreting culturally grounded expressions and the additional instruction-following limitations observed in the weakest evaluated model.

\begin{table}[!t]
\centering
\small
\begin{tabular}{lccc}
\toprule
\textbf{Error Type} & \textbf{GPT-4.1} & \textbf{Atlas-9B} & \textbf{Jais-590M} \\
\midrule
Correct              & 57 (53.3\%) & 34 (31.8\%) &  3 (2.7\%)\\
Literal Translation  & 43 (40.2\%) & 41 (38.3\%) & 18 (16.0\%) \\
Semantic Shift       & 7 (6.5\%) & 27 (25.2\%) & 18 (16.0\%) \\
Hallucination        & 0 (0.0\%) & 0 (0.0\%) & 59 (52.7\%) \\
Omission             & 0 (0.0\%) & 0 (0.0\%) & 1 (0.9\%) \\
Other                & 0 (0.0\%) & 5 (4.7\%) & 13 (11.6\%) \\
\bottomrule
\end{tabular}
\caption{Distribution of manually annotated categories across 100 translations per model. Values indicate counts and percentages of all assigned labels. A translation may receive multiple error labels.}
\label{error_analysis}
\end{table}

\subsection{Similarity-based Probing Analysis}
%\todo{EMM: we can moderate the claim, to explain that unsuccessful probing results may be due to weak crosslingual representation, not necessarily just figurative language understanding}

We conducted a similarity-based probing analysis to explore how the representations of Moroccan Darija sentences align with their English literal and figurative translations across model layers. We analyzed the 13 open-weight, decoder-only models evaluated in this study from the Atlas-Chat, AceGPT, JAIS, ALLaM, and Fanar families. The analysis measures whether Darija representations are closer to the intended or literal English translation. However, these scores may also be influenced by differences between Darija and English and by the wording and structure of the two translations.
Let
$$
\mathcal{D} = \{(x_j, y_j^{fig}, y_j^{lit})\}_{j=1}^{m}$$
%\end{equation}
denote a dataset of $m$ instances, where $x_j$, $y_j^{\text{fig}}$, and $y_j^{\text{lit}}$ represent the source sentence, figurative translation, and literal translation, respectively. Given a language model with $L$ no. of layers, the hidden representation of token $i$ at layer $l$ is denoted as:
$$h_i^{(l)} \in \mathbb{R}^{d}$$
$d$ is the hidden dimension. We applied mean pooling over all token embeddings to obtain a sentence-level embedding, as follows:
\begin{equation}
s^{(l)} = \frac{1}{N}\sum_{i=1}^{N} h_i^{(l)}
\end{equation}
$N$ is the number of tokens for each sentence. \\
For each layer $l$, the cosine similarity between the source sentence representation and the figurative translation representation is compared with the cosine similarity between the source sentence representation and the literal translation representation. If a model is able to represent the meaning of the source sentence correctly, we would expect the cosine of the source sentence with the figurative translation to be higher.
The accuracy of this inequality being true is then computed as:
\begin{equation}
\frac{1}{m}\sum_{j=1}^{m}\mathbf{1}\left(\cos(s_{x_j}^{(l)}, s_{y_j^{fig}}^{(l)}) > \cos(s_{x_j}^{(l)}, s_{y_j^{lit}}^{(l)})\right)
\end{equation}
Where $\mathbf{1}(\cdot)$ is the indicator function.
\begin{figure}[htp]
    \centering
    \includegraphics[width=1\linewidth,keepaspectratio]{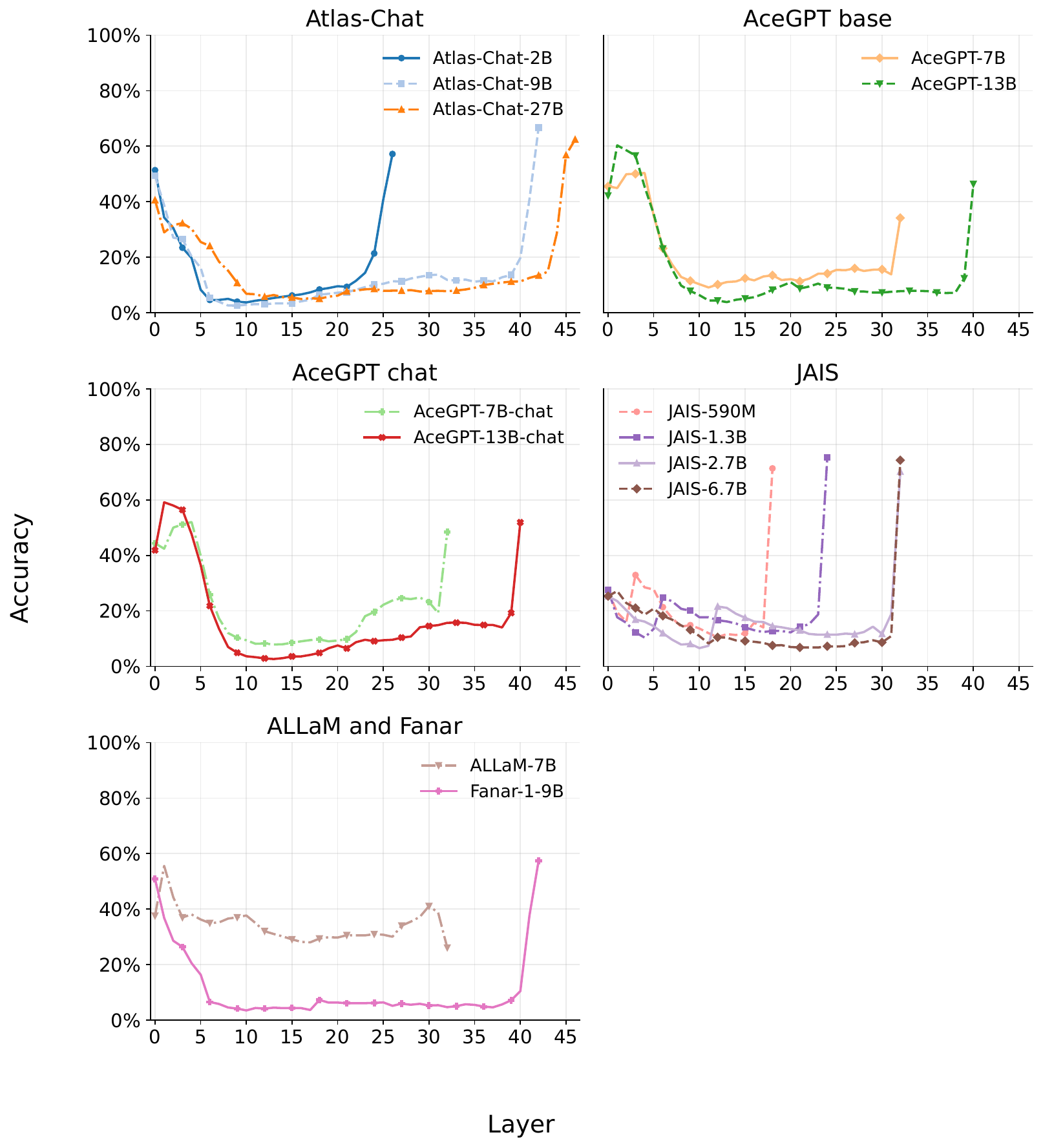}
\caption{Results for layer-wise similarity-based probing for figurative language understanding at context-level across layers.}
\label{analysis}
\end{figure}

Figure~\ref{analysis} shows the accuracy of context-level probing results by model families, where the length of each curve denotes different layer numbers. Increasing model size does not consistently ensure increased accuracy: Atlas-Chat-27B and both AceGPT-13B variants obtain higher scores in the early layers, but this advantage does not persist across intermediate layers. However, JAIS shows no clear analysis of model performance order based on model size.

Apart from increasing model size, layer choice also affects the rankings.While ALLaM shows a better accuracy in the middle layer, the output layer ranks the lowest among others (only around 26\%), whereas JAIS ranks around 70-75\%. However, the output layer accuracy does not necessarily reflect the translation and multiple-choice selection performance of each model. Although Atlas-Chat-9B achieves higher final-output probing accuracy and automatic translation scores than the 2B and 27B variants, Atlas-Chat-2B achieves higher MCQ accuracy. This mismatch is particularly apparent for JAIS-6.7B, which obtains high final-output probing accuracy despite low MCQ accuracy (14.88\%).
\begin{figure}[htp]
    \centering
    \includegraphics[width=\linewidth,height=16\textheight,keepaspectratio]{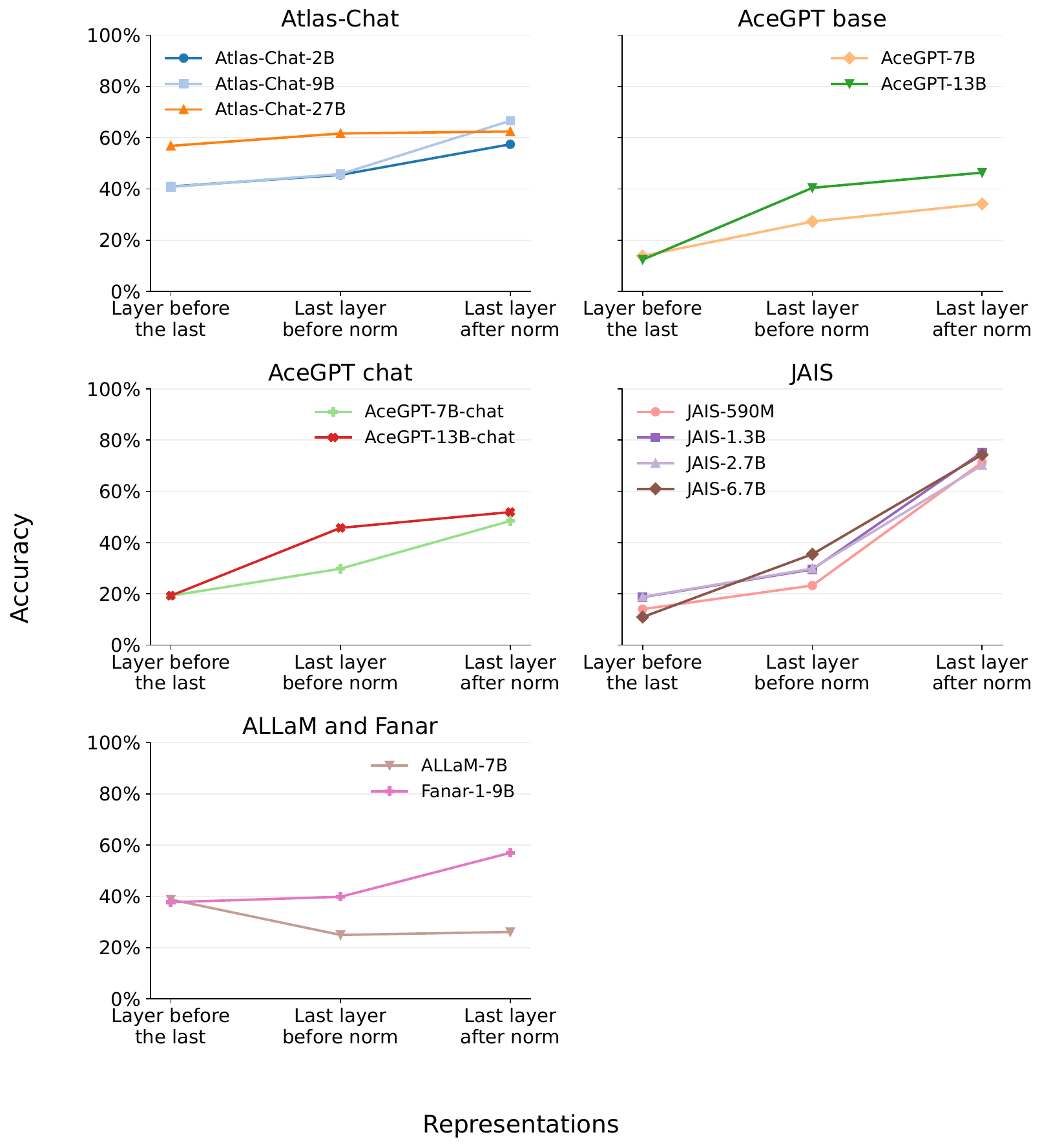}
\caption{Context-level similarity-based probing accuracy at the layer before the last, the last layer before final normalization, and the final output after normalization.}
\label{normalization}
\end{figure}
from final normalization, we compared representations from the layer before the last, the last layer before normalization, and the output after normalization (Figure~\ref{normalization}). JAIS and Fanar show large increases after normalization, whereas Atlas-Chat-27B and ALLaM change little at this step. ALLaM’s decrease occurs in the last layer before normalization. These results show that probing accuracy depends on where representations are taken within the model. In sum, this analysis provides further evidence on the representation similarities, subject to cross-lingual alignment constraints due to translation issues between Moroccan Darija and English along with effects of normalization and pooling on cosine similarity measurement.
Overall, the results show that even Arabic-specialized LLMs still face difficulties in forming reliable internal representations for figurative expressions, which makes their understanding especially challenging in low-resource settings, such as the case of Moroccan Darija.
\section{Conclusion}
\label{conclusion}
This work introduces MDFC, a benchmark for evaluating the interpretation and translation of figurative language in Moroccan Darija. Our experiments reveal differences in performance across models, with closed-weights systems generally better preserving intended meanings, while literal translation remains a recurring limitation even among high-performing models. Context and few-shot examples improve performance for some models, but their benefits vary across models and tasks. The agreement between human and LLM-based judgments further supports the usefulness of the LLM-as-a-judge for comparative evaluation within this benchmark. The error analysis highlights persistent difficulties in conveying figurative meanings, with lower-performing models also generating unrelated content or explaining and responding to expressions instead of translating them.

Future work will include expanding the benchmark to additional Arabic dialects and exploring approaches to improving figurative language understanding in LLMs.
\section*{Limitations}

This study has the following limitations: The dataset is relatively small and may not fully reflect the diversity of figurative expressions used in Moroccan Darija. The context-level dataset also contains only short contextual sentences, limiting the evaluation of figurative expressions within longer passages. Furthermore, the focus on Moroccan Darija means that the findings may not generalize to other Arabic dialects.

Another limitation concerns dataset construction. Although \underline{all generated samples} were reviewed and validated by native speakers, part of the construction process relied on LLM-assisted generation rather than entirely on naturally occurring conversational data. In particular, contextual sentences were created by conditioning the generation on the intended translation of each figurative expression. These sentences may therefore not fully capture the variety of natural contexts in which an expression can occur, especially the most idiosyncratic ones, or fail to capture the full spectrum of its nuances.

Finally, interpreting the probing results requires considering the cross-lingual nature of the task. Performance may reflect both figurative language understanding and the quality of cross-lingual representations between Moroccan Darija and English, making it difficult to separate the figurative understanding of the models from their general level of semantic competence of the language.

%\section*{Acknowledgments}

%Bibliography
\bibliographystyle{unsrt}  
\bibliography{references}  
\appendix
\section{Prompt Design}
\begin{figure*}[ht]
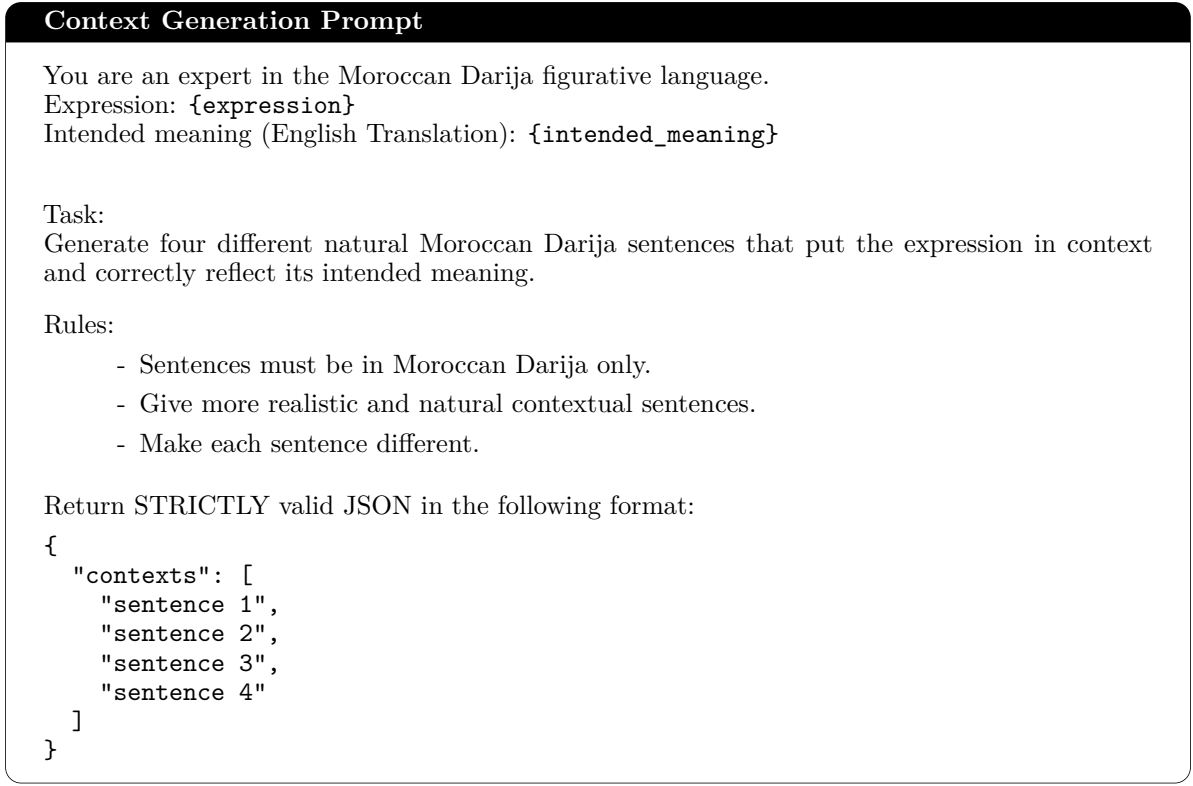

\centering
\begin{tcolorbox}[colback=white,
    colframe=black,
    width=0.95\linewidth,
    arc=2mm,
    boxrule=0.3pt,
    title=\textbf{Context Generation Prompt}
]
You are an expert in the Moroccan Darija figurative language.\\
Expression: \texttt{\{expression\}}\\
Intended meaning (English Translation): \texttt{\{intended\_meaning\}}\\
\vspace{0.3cm}\\
Task:\\
Generate four different natural Moroccan Darija sentences that put the expression in context and correctly reflect its intended meaning.
\vspace{0.3cm}\\
Rules:
\begin{itemize}
    \item[-] Sentences must be in Moroccan Darija only.
    \item[-] Give more realistic and natural contextual sentences.
    \item[-] Make each sentence different.
\end{itemize}
\vspace{0.3cm}
Return STRICTLY valid JSON in the following format:
\begin{verbatim}
{
  "contexts": [
    "sentence 1",
    "sentence 2",
    "sentence 3",
    "sentence 4"
  ]
}
\end{verbatim}
\end{tcolorbox}
\caption{Prompt used for generating contextual Moroccan Darija sentences containing figurative expressions, where \texttt{\{expression}\} and \texttt{\{intended\_meaning}\} represent the source expression and its intended figurative English translation, respectively.}
\label{prompt_template1}
\end{figure*}

\begin{figure*}[ht]
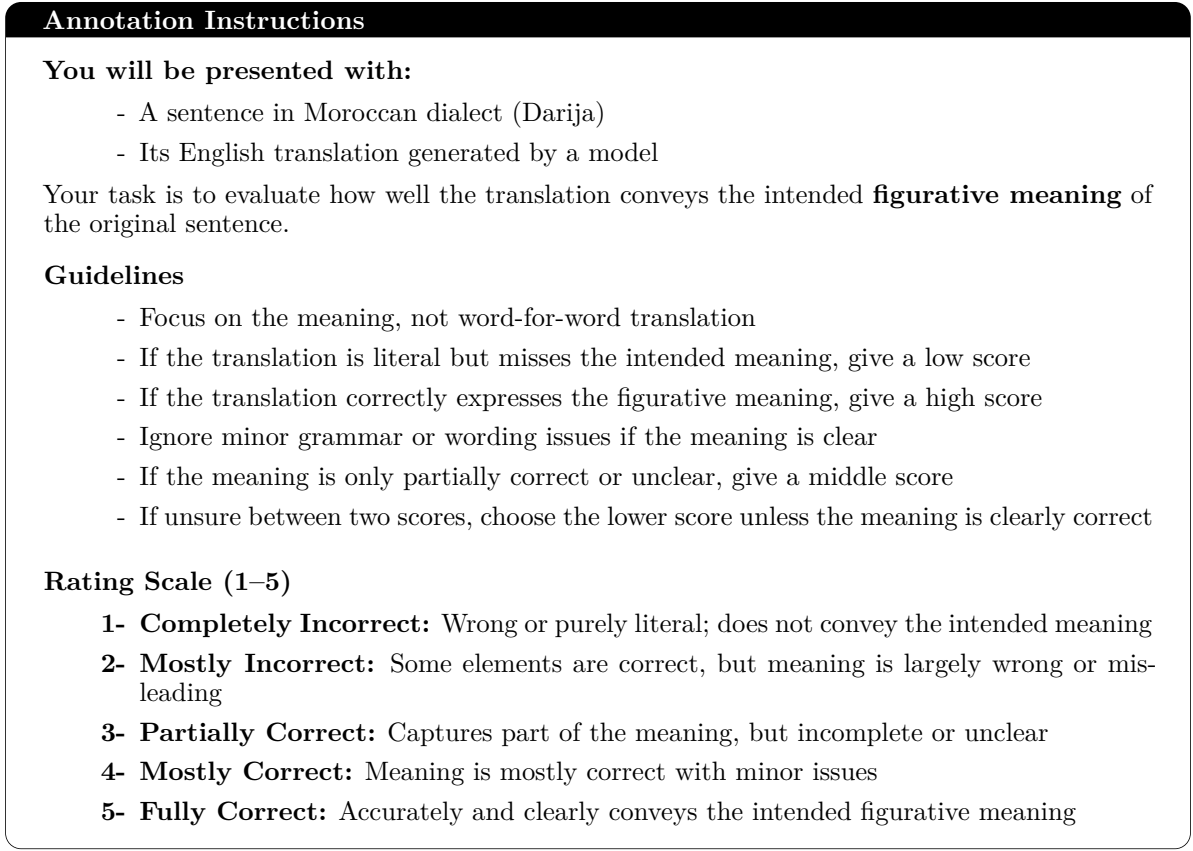


\centering

\begin{tcolorbox}[colback=white,
    colframe=black,
    width=0.95\linewidth,
    arc=2mm,
    boxrule=0.3pt,
    title=\textbf{Annotation Instructions}
]

\textbf{You will be presented with:}

\begin{itemize}

    \item[-] A sentence in Moroccan dialect (Darija)
    \item[-] Its English translation generated by a model

\end{itemize}

Your task is to evaluate how well the translation conveys the intended \textbf{figurative meaning} of the original sentence.

\vspace{0.3cm}

\textbf{Guidelines}

\begin{itemize}

    \item[-] Focus on the meaning, not word-for-word translation
    \item[-] If the translation is literal but misses the intended meaning, give a low score
    \item[-] If the translation correctly expresses the figurative meaning, give a high score
    \item[-] Ignore minor grammar or wording issues if the meaning is clear
    \item[-] If the meaning is only partially correct or unclear, give a middle score
    \item[-] If unsure between two scores, choose the lower score unless the meaning is clearly correct

\end{itemize}

\vspace{0.3cm}

\textbf{Rating Scale (1--5)}

\begin{enumerate}[label=\textbf{\arabic*-}]

    \item \textbf{Completely Incorrect:} Wrong or purely literal; does not convey the intended meaning
    \item \textbf{Mostly Incorrect:} Some elements are correct, but meaning is largely wrong or misleading
    \item \textbf{Partially Correct:} Captures part of the meaning, but incomplete or unclear
    \item \textbf{Mostly Correct:} Meaning is mostly correct with minor issues
    \item \textbf{Fully Correct:} Accurately and clearly conveys the intended figurative meaning

\end{enumerate}

\end{tcolorbox}

\caption{Human evaluation guidelines for figurative translation assessment.}
\label{prompt2}
\end{figure*}

This section presents the context generation prompt used to produce Moroccan Darija sentences containing figurative expressions and reflecting their intended meanings (Figure~\ref{prompt_template1}), as well as the human evaluation guidelines for assessing model performance in translating figurative expressions (Figure~\ref{prompt2}).

\section{Statistical Significance Analysis}
\label{SA}
\begin{figure}[t]
    \centering
    \begin{minipage}{0.48\textwidth}
        \centering
        \includegraphics[width=\linewidth,height=0.4\textheight]{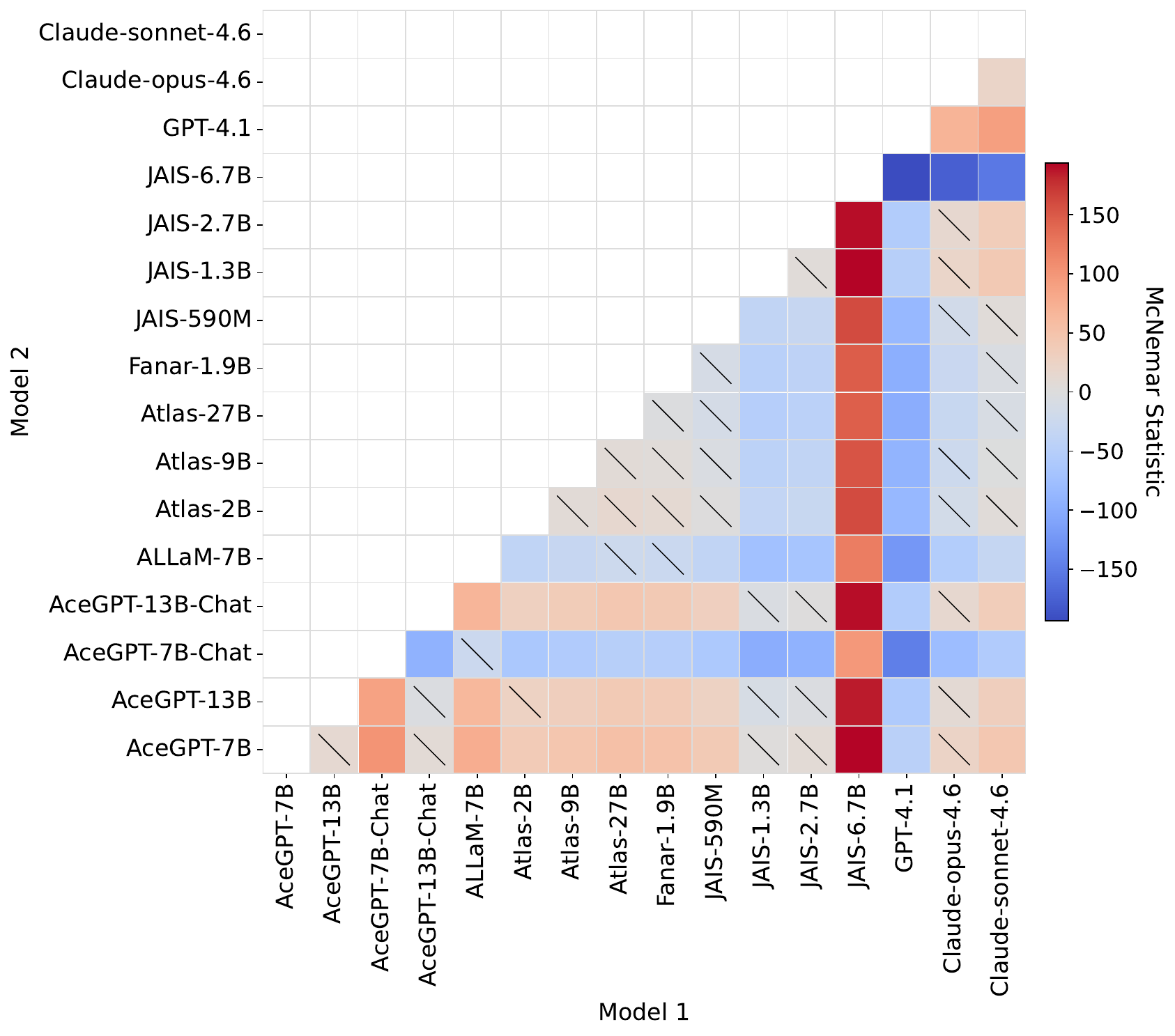}
        \subcaption{(a)} 
    \end{minipage}
    \hfill
    \begin{minipage}{0.48\textwidth}
        \centering
        \includegraphics[width=\linewidth,height=0.4\textheight]{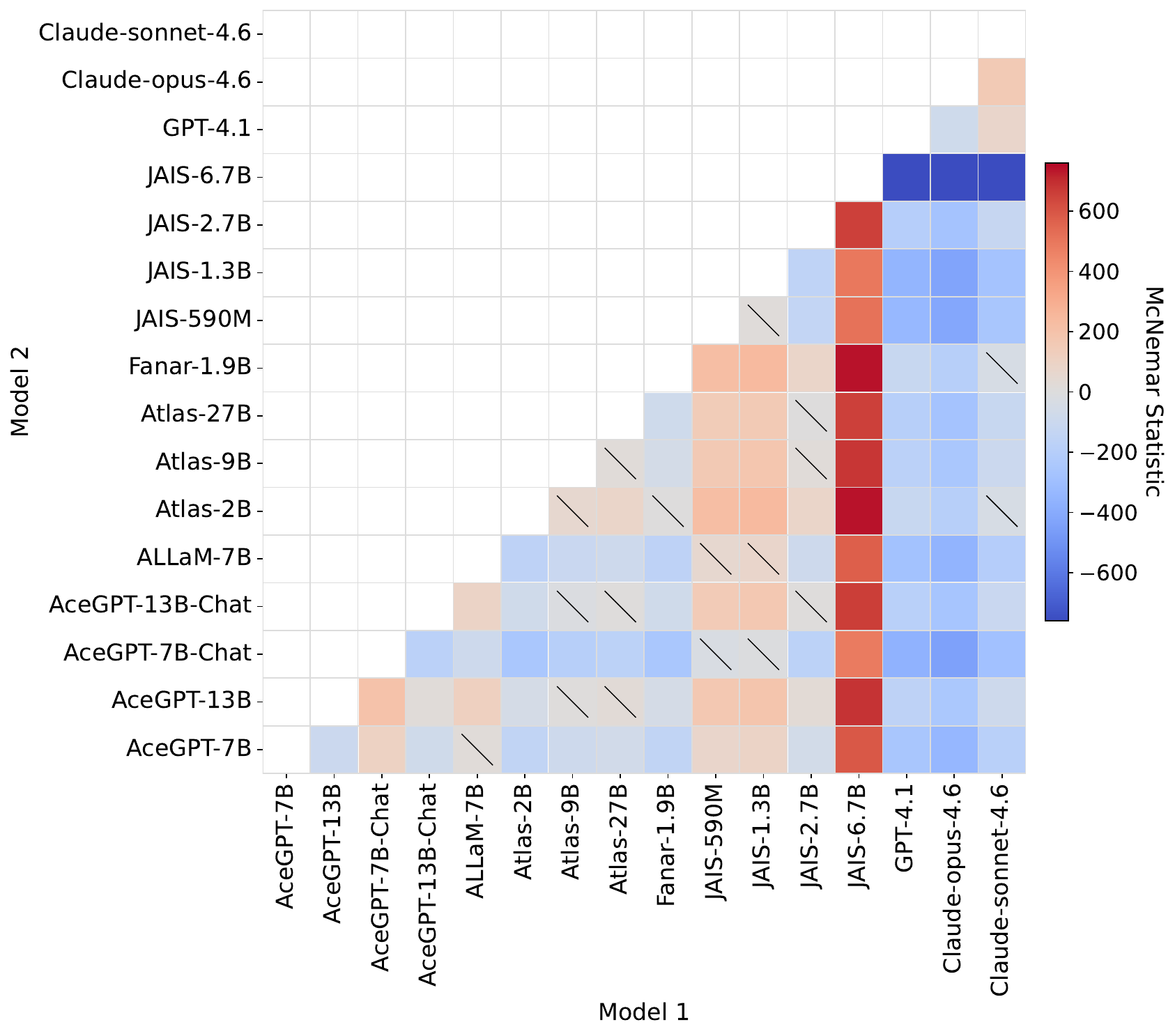}
        \subcaption{(b)} 
    \end{minipage}
    \caption{Pairwise McNemar test results for the zero-shot setting on the (a) expression-level and (b) context-level tasks. Each cell compares Model 1 on the x-axis with Model 2 on the y-axis. Red indicates that Model 2 achieves higher accuracy than Model 1, whereas blue indicates the opposite. Cells marked with a diagonal slash denote non-significant differences after Benjamini–Hochberg correction.}
\label{mcnemar}
\end{figure}

To examine whether the difference in model performance on MCQ tasks is statistically significant, we carried out pairwise McNemar tests~\citep{McNemar1947} in the zero-shot scenario. The p-values obtained from these pairwise McNemar tests were adjusted using the Benjamini-Hochberg correction~\citep{benjamini}.  As depicted in Figure~\ref{mcnemar}, majority of model pair comparisons provide statistically significant p-values, implying that the observed performance differences are robust and unlikely to have occurred by chance. In particular, the top-performing models (closed-weights): Claude-opus-4.6, Claude-sonnet-4.6, and GPT-4.1 perform better compared to other open-weights models. The analysis shows no statistical significance between models whose performances are nearly equivalent and/or belong to the same model group/parameter size category. Moreover, the context-level benchmark displays more statistically significant differences between model comparisons in contrast to the expression level, suggesting that contextual information provides stronger discrimination between model performance.

\begin{figure}[!htp]

    \centering
    \begin{minipage}{0.48\textwidth}
        \centering
        \includegraphics[width=\linewidth,height=0.4\textheight]{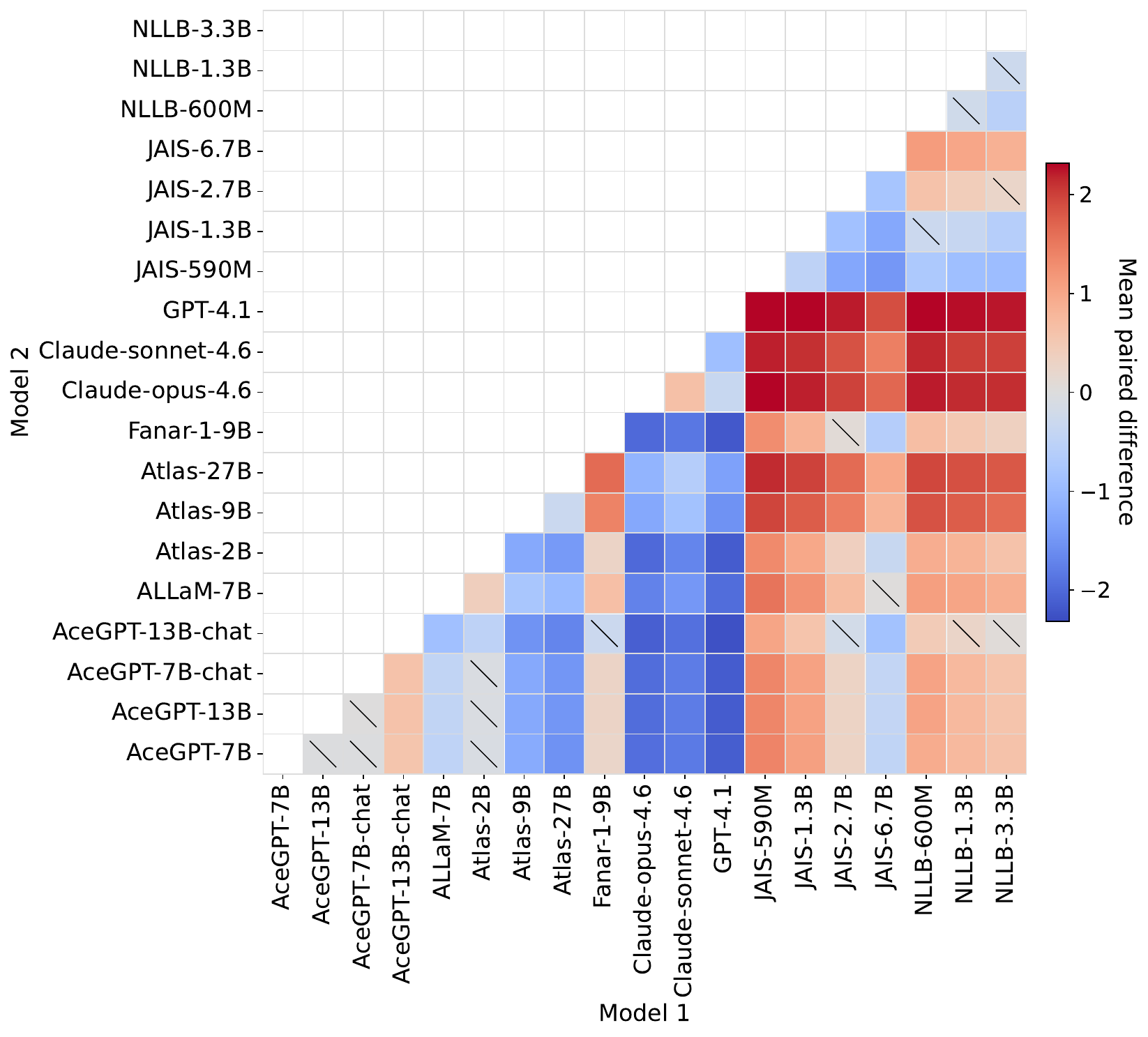}
        \subcaption{(a)} 
    \end{minipage}
    \hfill
    \begin{minipage}{0.48\textwidth}
        \centering
        \includegraphics[width=\linewidth,height=0.4\textheight]{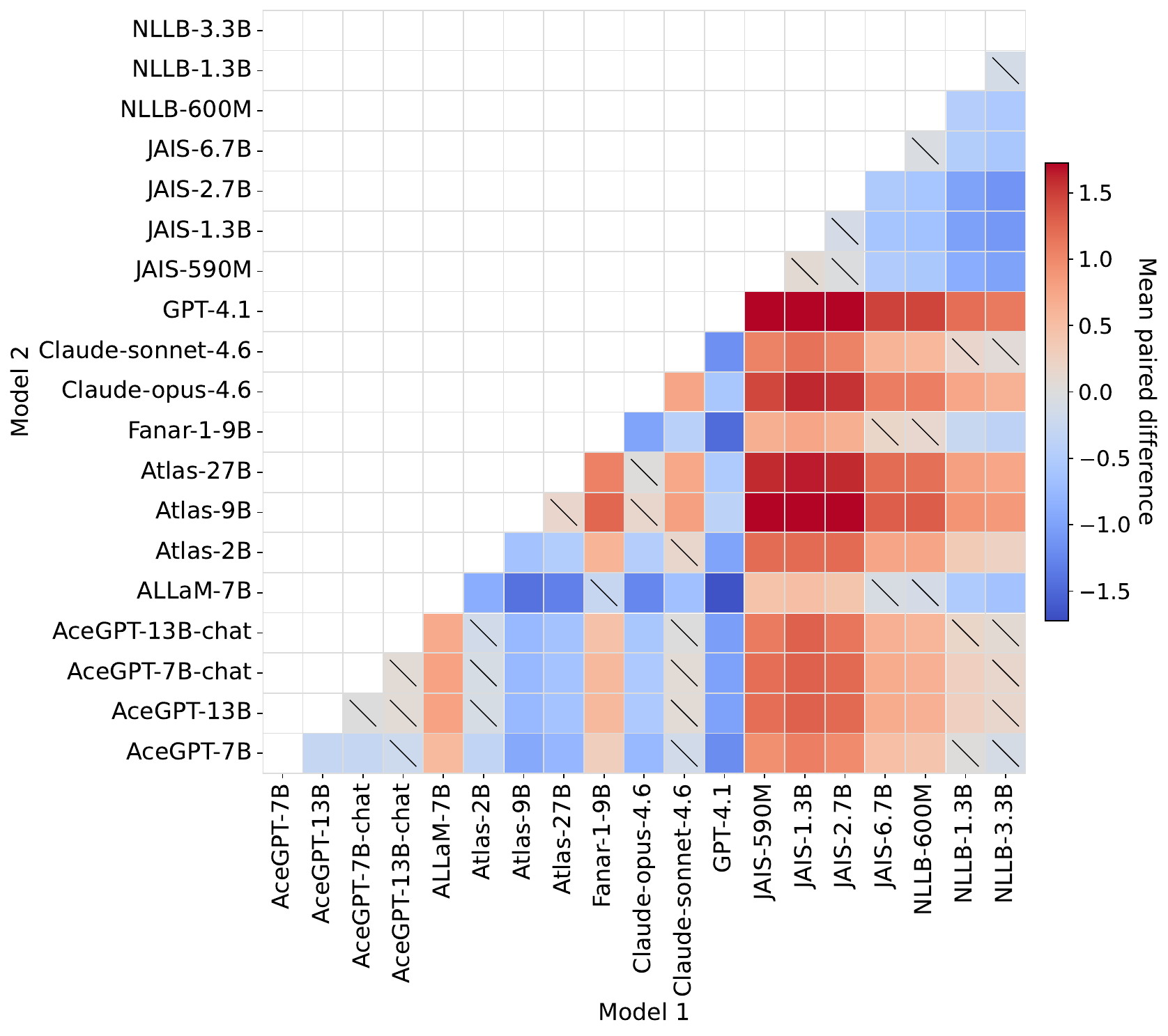}
        \subcaption{(b)} 
    \end{minipage}
    \hfill
    \caption{Pairwise Wilcoxon signed-rank test results for the expression-level LLM-as-a-judge evaluation. (a) Adequacy and (b) Fluency. Colored cells indicate statistically significant differences between model pairs ($p < 0.05$). Red cells denote that Model 2 received significantly higher scores than Model 1. Color intensity reflects the magnitude of the mean paired-score difference, with darker shades indicating larger differences. Cells marked with a diagonal dash represent non-significant differences.}
    \label{wil_exp}
\end{figure}

\begin{figure}[!htp]
    \centering
    \begin{minipage}{0.48\textwidth}
        \centering
        \includegraphics[width=\linewidth,height=0.4\textheight]{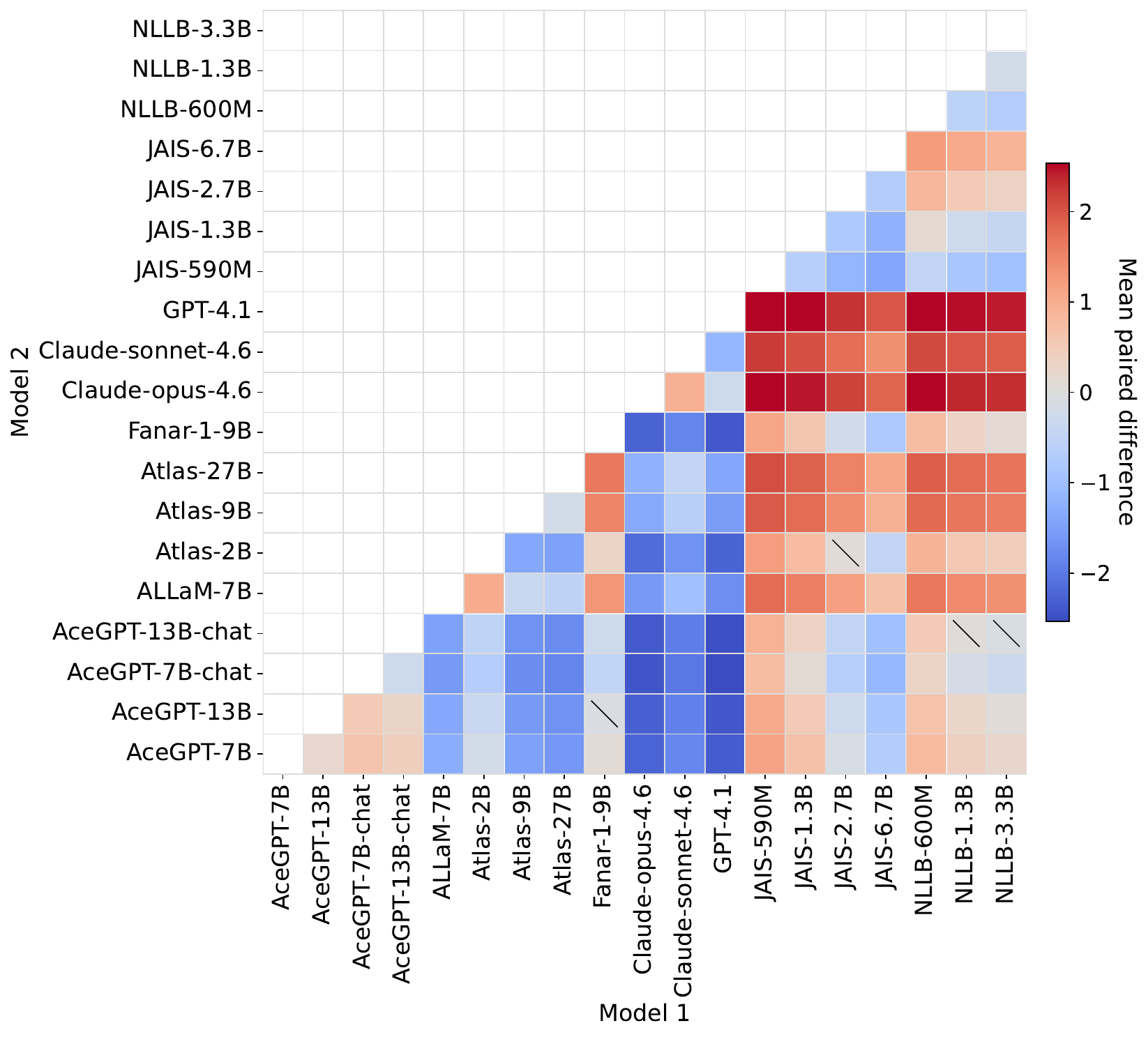}
        \subcaption{(a)} 
    \end{minipage}
    \hfill
    \begin{minipage}{0.48\textwidth}
        \centering
        \includegraphics[width=\linewidth,height=0.4\textheight]{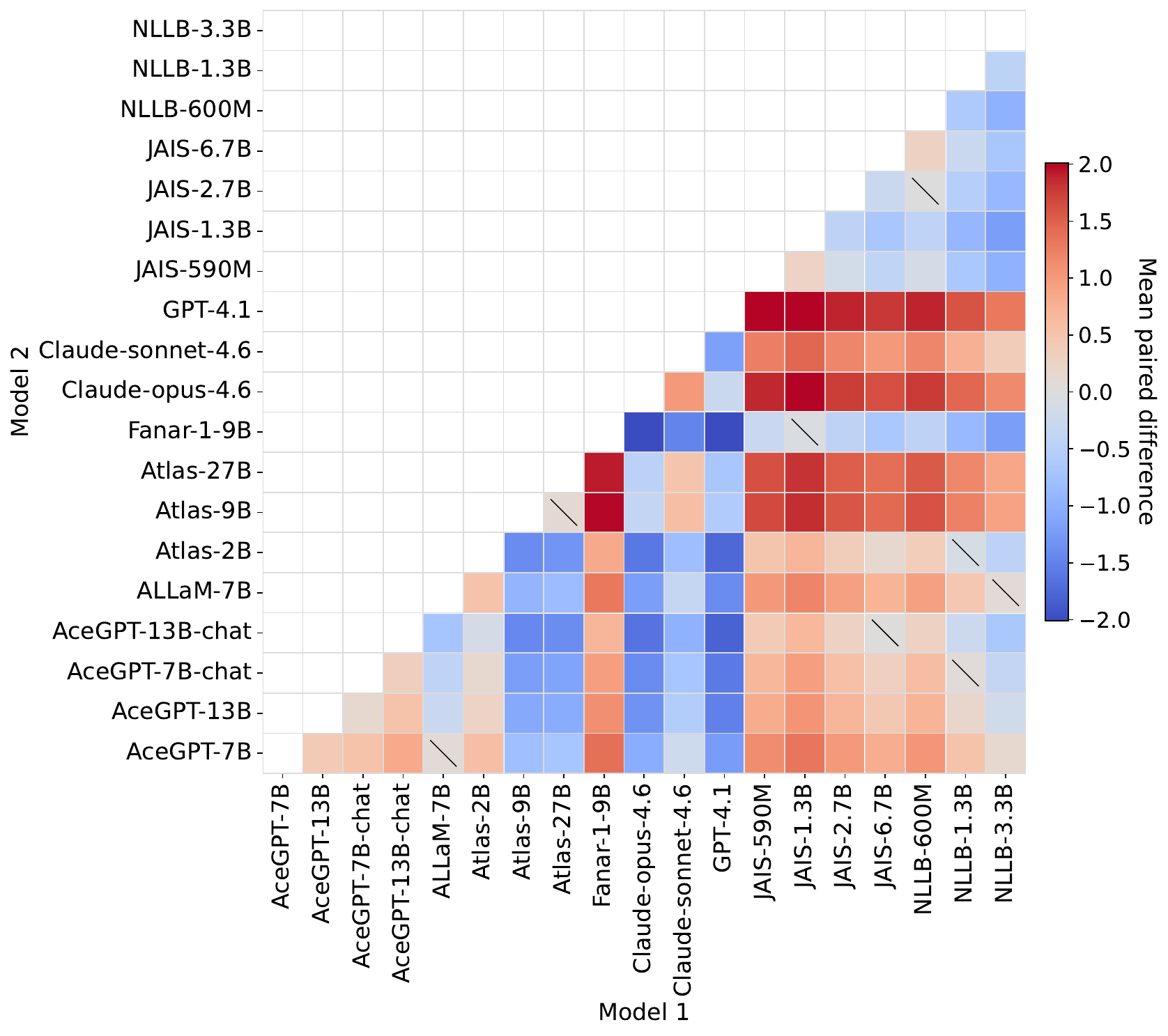}
        \subcaption{(b)} 
    \end{minipage}
    \hfill
    
    \caption{Pairwise Wilcoxon signed-rank test results for the context-level LLM-as-a-judge evaluation. (a) Adequacy and (b) Fluency. Colored cells indicate statistically significant differences between model pairs ($p < 0.05$). Red cells denote that Model 2 received significantly higher scores than Model 1. Color intensity reflects the magnitude of the mean paired-score difference, with darker shades indicating larger differences. Cells marked with a diagonal dash represent non-significant differences.}
    \label{wil_ctx}
\end{figure}

For the LLM-as-a-judge evaluation, we performed pairwise Wilcoxon signed-rank~\citep{Rey2011} tests on the adequacy and fluency scores. These p-value results were subsequently adjusted with Benjamini-Hochberg method~\citep{benjamini}. According to Figures~\ref{wil_exp} and~\ref{wil_ctx}, statistically significant pairwise differences were mostly obtained on both the expression-level and context-level benchmarks. Claude-opus-4.6, Claude-sonnet-4.6, and GPT-4.1 receive significantly higher adequacy and fluency ratings than most open-source models. Non-significant differences occur both within and across model families, particularly for fluency on the expression-level benchmark. The context-level benchmark shows fewer non-significant comparisons across both dimensions, indicating more distinguishable model performance when expressions are presented in context.

\end{document}